\documentclass[10pt,twocolumn,letterpaper]{article}

\usepackage{algorithm}

\usepackage[pagenumbers]{cvpr} 

\usepackage{algorithmic}
\usepackage[dvipsnames]{xcolor}

\usepackage{multirow}
\definecolor{cvprblue}{rgb}{0.21,0.49,0.74}
\usepackage[pagebackref,breaklinks,colorlinks,citecolor=cvprblue]{hyperref}

\def\paperID{38} 
\def\confName{CVPR}
\def\confYear{2024}

\title{Data-free Defense of Black Box Models Against Adversarial Attacks}
\author{Gaurav Kumar Nayak \\ \textit{University of Central Florida} \and Inder Khatri \\ \textit{New York University} \and Ruchit Rawal \hspace{5pt} Anirban Chakraborty \\ \textit{Indian Institute of Science}}

\begin{document}
\maketitle
\begin{abstract}
Several companies often safeguard their trained deep models (i.e. details of architecture, learnt weights, training details etc.) from third-party users by exposing them only as black boxes through APIs. Moreover, they may not even provide access to the training data due to proprietary reasons or sensitivity concerns. In this work, we propose a novel defense mechanism for black box models against adversarial attacks in a data-free set up. We construct synthetic data via generative model and train surrogate network using model stealing techniques. To minimize adversarial contamination on perturbed samples, we propose `wavelet noise remover' (WNR) that performs discrete wavelet decomposition on input images and carefully select only a few important coefficients determined by our `wavelet coefficient selection module' (WCSM). To recover the high-frequency content of the image after noise removal via WNR, we further train a `regenerator' network with an objective to retrieve the coefficients such that the reconstructed image yields similar to original predictions on the surrogate model. At test time, WNR combined with trained regenerator network is prepended to the black box network, resulting in a high boost in adversarial accuracy. Our method improves the adversarial accuracy on CIFAR-$10$ by $38.98\%$ and $32.01\%$ on state-of-the-art Auto Attack compared to baseline, even when the attacker uses surrogate architecture (Alexnet-half and Alexnet) similar to the black box architecture (Alexnet) with same model stealing strategy as defender. The code is available at \href{https://github.com/vcl-iisc/data-free-black-box-defense}{https://github.com/vcl-iisc/data-free-black-box-defense}
\end{abstract}
\vspace{-10pt}
\section{Introduction}
\label{sec:intro}
Deep neural networks, applied in computer vision~\cite{Voulodimos2018DeepLF,Rawat2017DeepCN}, machine translation~\cite{Bahdanau2015NeuralMT,mi2022improving}, speech recognition~\cite{lei2022bat,fayek2017evaluating}, have exhibited success but face unreliability due to adversarial attacks causing erroneous predictions~\cite{akhtar2018threat,huang2022defeat,Szegedy2014IntriguingPO,vidnerova2020vulnerability}. These attacks can be categorized into either black box~\cite{papernot2017practical,Chen2017ZOOZO,zhao2023remix,hao2022boosting} and white box~\cite{goodfellow2014explaining,Madry2018TowardsDL,Kurakin2017AdversarialEI,Croce2020ReliableEO} attacks based on access to model parameters. Black box attacks, more practical and realistic than white-box attacks, involve stealing the functionality of target models by training surrogate models using pairs of (image, predictions). Adversarial samples crafted using surrogate models can also exploit the property of transferability to attack the target model. Hence, immediate attention is required to protect against such attacks.

To make it harder for the adversary to craft black box attacks, companies prefer not to release the training dataset and keep them proprietary. However, recent works have shown that model stealing can compromise the confidentiality of black-box models even without the training data. Existing works perform generative modeling either with proxy data~\cite{orekondy2019knockoff,barbalau2020black,sanyal2022towards} or without proxy data~\cite{kariyappa2021maze,truong2021data,zhou2020dast} and train the surrogate model with the synthesized data for model stealing. However, their focus is more on obtaining highly accurate surrogate models. In contrast to existing works, we inquire about an important question regarding the safety of the black box models - “{\em how to defend against black box attacks in data-free (absence of training data) set up}”.  

\begin{figure}[htp]
\centering
\centerline{\includegraphics[width=0.5\textwidth]{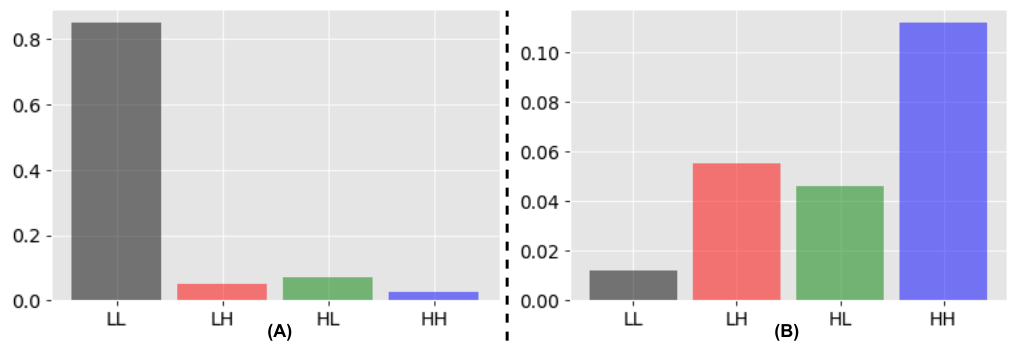}}
\vspace{-6pt}
\caption{The average absolute magnitude of approximate (LL) and detail coefficients (LH, HL and HH) (via 
wavelet decomposition) across samples on a) clean data and b) normalized difference between wavelet decomposition of clean and corresponding adversarial image. In (a) the lesser contaminated LL coefficients have higher magnitude. 
In (b) LL are least affected. 
}
\label{fig:wavelet_bargraph}
\end{figure}

In order to tackle this problem, our proposed method `\textit{DBMA}' (i.e., \textbf{d}efending \textbf{b}lack box \textbf{m}odels against \textbf{a}dversarial attacks in data-free setup) leverages on the wavelet transforms~\cite{waveletssurvey}. 
We observe difference between wavelet transform on adversarial sample and original sample (shown in Fig.~\ref{fig:wavelet_bargraph} (B))
, we notice that detail coefficients in high-frequency regions (LH, HL and HH regions) are majorly corrupted by adversarial attacks and the approximate coefficients (LL region) is least affected for level 1 decomposition. Similar observation holds even for decomposition on other levels. To improve the adversarial accuracy, a naive way would be to completely remove the detail coefficients which can minimize the contamination in adversarial samples but it can lead to a huge drop in clean accuracy as the model predictions are highly correlated with high frequencies~\cite{wang2020high}. To avoid that, we assign importance to each of the detail coefficients based on magnitude. The least perturbed LL region usually contains higher magnitude coefficients than other regions (Fig.~\ref{fig:wavelet_bargraph} (A)). So, we prefer high magnitude detail coefficients. However, taking a large number of such coefficients can lead to good clean accuracy but at the cost of low adversarial accuracy due to more contamination. On the other hand, taking very few such coefficients can allow lower contamination but results in low clean accuracy. Our method judiciously takes care of this trade-off, and carefully selects the required important detail coefficients (discussed in Sec.~\ref{subsec:WCSM}) using the proposed wavelet coefficient selection module (WCSM).
\vspace{-8pt}

The wavelet noise remover (WNR) removes noise coefficients by filtering out only the top-$k$\% high magnitude coefficients where optimal $k$ is selected using the WCSM module. As a side-effect, a lot of high-frequency content of the image gets lost which reduces the overall discriminability and ultimately results in suboptimal model’s performance. To cope up with this reduced discriminability, we introduce a U-net-based regenerator network(Sec.~\ref{subsec:Regenerator_network}), that takes the spatial samples corresponding to selected coefficients as input and outputs the reconstructed image. 
The regenerator network is trained  by regularizing the feature and input space of the reconstructed image on the surrogate model. In the feature space, we apply cosine similarity and kl-divergence losses to ensure proper reconstruction on clean and adversarial data respectively. Besides this, we also regularize the input space using our spatial consistency loss. Finally, the WNR module combined with the trained regenerator network is appended before the black-box model. 
The attacker has black-box access to the complete end to end model containing the defense components.

We summarize our contributions as follows:
\begin{itemize}
\item 
In this work, we investigate the largely underexplored yet critical challenge of defending against adversarial attacks on a \textit{black box model without access to the network weights and in the absence of original training samples}. 
\item We propose a novel strategy to provide adversarial robustness against data-free black box attacks by introducing two key defense components: 
\newline
    i.) We propose a wavelet-based noise remover (WNR) containing selective wavelet coefficient module (Sec.~\ref{subsec:WCSM}) that aims to remove coefficients corresponding to high frequency components which are most likely to be corrupted by adversarial attack.
\newline
    ii.) We propose a U-net-based regenerator network (Sec.~\ref{subsec:Regenerator_network}) that retrieves the 
coefficients that are lost after the noise removal (via WNR) so that the high-frequency image content can be restored.
\item We demonstrate the efficacy of our method via extensive experiments and ablations on both the components of proposed framework namely wavelet noise remover (Sec.~\ref{subsec:coefficients_quantity},~\ref{subsec:benefit_WNR}) and the regenerator network (Sec.~\ref{subsec:losses}) which are appended before the black box target model. The resulting combined model used as the new black box (as seen by attacker), yields high clean and adversarial accuracy on test data (Sec.~\ref{subsec:datasets}).
\end{itemize}
\section{Related Works}
\label{sec:related}
Our work is closely related to model stealing and wavelets, so we briefly discuss their related works below.

\textbf{Data-efficient Model stealing}: Based on the availability of training data, we categorize model stealing works as follows:\\
\textbf{Training data} - On full training data, knowledge distillation~\cite{hinton2015distilling} is used to extract knowledge using soft labels obtained from the black box model. With few training samples, Papernot~\textit{et al.}~\cite{papernot2017practical} generates additional synthetic data in the directions (computed using jacobian) where model’s output varies in the neighborhood of training samples. \\
\textbf{Proxy data} - In the absence of training data, either natural or synthetic images are used as proxy data. Orekondy~\textit{et al.}~\cite{orekondy2019knockoff} query the black box model on the natural images using adaptive strategy via reinforcement learning to get output predictions and use them to replicate the functionality of the black box model. Barbalau~\textit{et al.}~\cite{barbalau2020black} use evolutionary framework to learn image generation on a proxy dataset where the generated images are enforced to exhibit high confidence on the black box model. Sanyal~\textit{et al.}~\cite{sanyal2022towards} use the GAN framework with a proxy dataset composed of either related/unrelated data or synthetic data.\\
\textbf{Without Proxy data} - Kariyappa~\textit{et al.}~\cite{kariyappa2021maze} proposed an alternate training mechanism between generator and surrogate model, where generator is trained to produce synthetic samples to maximize the discrepancy between the predictions of the surrogate and the black box model. Truong~\textit{et al.}~\cite{truong2021data} also train generator and surrogate model alternatively but they replace discrepancy loss computed using KL divergence with L1 norm over logits approximated from softmax. Similarly, Zhou~\textit{et al.}~\cite{zhou2020dast} also formulate a min-max adversarial game 
but they additionally enforce the synthetic data to be equally distributed among the classes using a conditional generator.

Unlike these existing works, we use model stealing only as a means to obtain the surrogate model and synthetic data. Unlike them where they steal as an adversary, our goal is to provide robustness against black box attacks i.e. to reduce the effects of the adversarial samples on the black box model that are crafted using the surrogate model.

\textbf{Wavelet in CNNs}: Before the CNNs, the wavelets had been used for noise reduction and denoising~\cite{donoho1994ideal,donoho1995noising}. 
Prakash~\textit{et al.}~\cite{prakash2018deflecting} used pixel deflection technique followed by adaptive thresholding on the wavelets coefficients for denoising. Unlike these works, our setup is more challenging due to no access to both the training data and the model weights. Mustafa~\textit{et al.}~\cite{mustafa2019image} utilized wavelet denoising with image superresolution as a defense mechanism against adversarial attacks in grey-box settings. 
Different from this work, our approach utilizes the wavelet with a proposed regenerator network for defense against adversarial attacks in a black-box setting.

\textbf{Data and Training Efficient Adversarial Defense:} Adversarial Defense techniques can be broadly classified into two categories: Adversarial Training (AT) and Non-Adversarial Training (Non-AT) based methods. AT-based methods  \citep{goodfellow2014explaining, Madry2018TowardsDL,LAMB2022218,chen2022towards} rely on adversarial samples during training to improve performance on  perturbed samples. However, these methods are computationally expensive and often require high-capacity networks for achieving significant gains in robustness \cite{Zi_2021_ICCV}. On the other hand, Non-AT-based methods like JARN\cite{chan2019jacobian}, BPFC\cite{Addepalli_2020_CVPR}, and GCE\cite{chen2019improving} offer faster training but perform inadequately against a wide range of strong attacks \cite{heattack2017}. 

In recent years, researchers have proposed adversarial defense methods that do not require re-training a model for providing robustness, thus making them train efficiently. This has immense practical benefits as unlike most state-of-the-art defenses that necessitate model re-training, such approaches can be seamlessly integrated with already deployed models. One such method, namely Magnet\cite{meng2017magnet}, first classifies inputs as clean or adversarial and then transforms the adversarial inputs closer to the clean image manifold. Another approach, Defense-GAN by Samangouei et al.\cite{samangouei2018defense}, uses Generative Adversarial Networks to learn the distribution of clean images and generate samples similar to clean images from inputs corrupted with adversarial noise. Sun et al.\cite{sun2019adversarial} introduced the Sparse Transformation layer (STL) which maps input images to a low-dimensional quasi-natural image space, suppressing adversarial contamination and making adversarial and clean images indistinguishable. Theagarajan et al.\cite{Theagarajan_2020_CVPR_Workshops} presented a defense protocol for black-box facial recognition classifiers consisting of a Bayesian CNN-based adversarial attack detector and image purifiers trained using the data from similar domain to the original training data. They used ensemble of image purifiers for removing the adversarial noise and attack detector for validating the purified image. 
However, a key limitation of these methods is their reliance on the original training-data for training the defense components, which hinders their use in scenarios where training-data/statistics are not freely available due to proprietary to privacy reasons, etc. 


Consequently, recent advancements have redirected focus towards tackling the limitations of training data dependency in defending neural networks against adversarial attacks. For instance, Mustafa et al.\cite{mustafa2019image} provide defense on pretrained network without retraining or accessing training data, but have additional dependency on pretrained image super-resolution networks. Moreover, their 
approach can only be integrated with the target networks that are capable of handling multi-scale inputs, thus making them infeasible on the pretrained networks incapable of handling multi-scale inputs. To avoid these problems, Qiu et al.\cite{qiu2021efficient} proposed the RDG (Random Distortion over Grids) preprocessing operation, randomly distorting input images by dropping and displacing pixels. Guesmi et al.\cite{guesmi2021sit} presented SIT (Stochastic Input Transformation), applying random transformations to eliminate adversarial perturbations while maintaining similarity to the original clean images. However, the stochastic nature of SIT without guidance negatively affects the clean accuracy of the target model. In contrast, our proposed Data-free Black-Box defense method, DBMA, better preserves clean-accuracy while also achieving higher robustness.

\section{Preliminaries}
\label{sec:prelims}

\textbf{Notations}: The black box model is denoted by $B_{m}$ which is trained on the proprietary training dataset $O_{d}^{train}$. We denote the surrogate model by $S_{m}$. The generator $G$ produces synthetic data $S_{d} = \{x_s^i\}_{i=1}^{N}$ containing $N$ samples. The logit obtained by the model $S_{m}$ on input $x$ is $S_{m}(x)$. The softmax and the label predictions on sample $x$ by model $S_{m}$ are represented by $soft(S_{m}(x))$ and $label(S_{m}(x))$.

The set $A_{a} = \{A_{a}^{p}\}_{p=1}^{P}$ contains $P$ different adversarial attacks. The adversarial sample corresponding to the $n^{th}$ sample of test dataset $O_{d}^{test}$  (i.e. $x_o^n$) is denoted by  $x_{oa}^n$ which is crafted with a goal to fool the network $B_{m}$. Similarly, $S_{da}$ is the set of crafted adversarial samples corresponding to the synthetic data $S_{d}$ where the adversarial sample $x_{sa}^{n} \in S_{da}$ is obtained by perturbing the synthetic sample $x_s^n \in S_{d}$ using an adversarial attack $A_a^{j} \in A_{a}$.
 
We denote the discrete wavelet transform and its inverse operation by $DWT(.)$ and $IDWT(.)$ respectively. The wavelet coefficient selection module is denoted by WCSM. The regenerator network is represented by $R_{n}$. 

\textbf{Model Stealing}: Model stealing involves extracting the black box model's ($B_{m}$) functionality by inputting images into its API to gather outputs, used subsequently to train a surrogate model ($S_m$). When no training data $O_{d}^{train}$ is accessible, this scenario is termed data-free model stealing.

\begin{figure*}[ht]
\centering
\centerline{\includegraphics[width=0.9\textwidth]{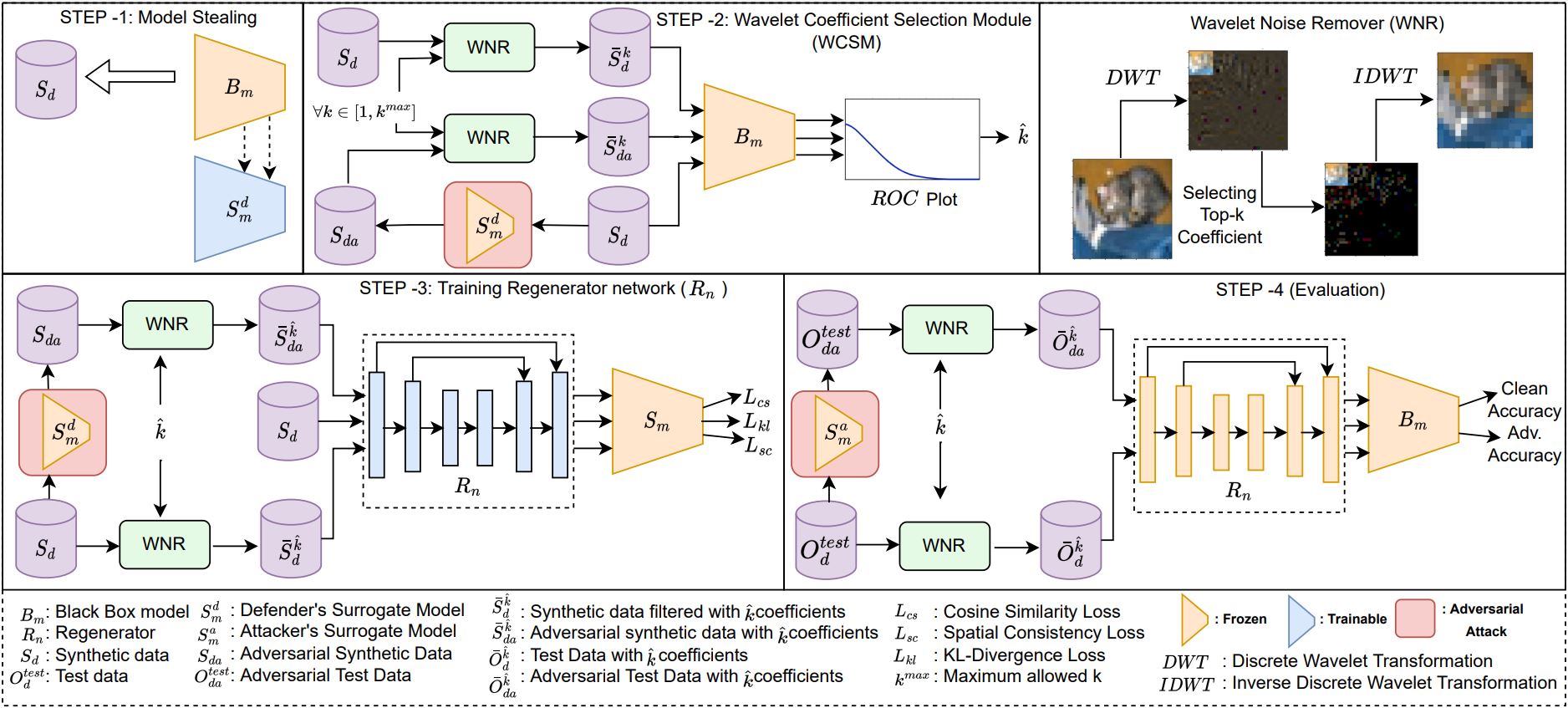}}
\vspace{-10pt}
\caption{\small{An overview of our proposed approach DBMA. In step $1$, we obtain the defender's surrogate model $S^d_m$ and synthetic data $S_d$ by model stealing from the victim model $B_m$. In step $2$, we use the Wavelet Coefficient Selection Module (WCSM) that gives the optimal \% of coefficients ($\hat{k}$) to be selected by the Wavelet Noise Remover (WNR) which are likely to be least corrupted by adversarial attacks. In step $3$, we train a regenerator network $R_{n}$ using different losses ($L_{cs}$, $L_{kl}$, $L_{sc}$) such that the model $S^d_m$ yields features on the regenerated data (clean $R_n(\bar{S}_{d}^{k})$ and adversarial $R_n(\bar{S}_{da}^{k})$) similar to the features on clean data $S_d$. 
Finally in step $4$, we evaluate our DBMA approach on test clean ($O_{d}^{test}$) and adversarial samples ($O_{da}^{test}$) where the WNR (with $k=\hat{k}$) and trained $R_n$ are prepended to $B_m$.}}
\label{fig:approach}
\end{figure*}
\textbf{Adversarial Attacks}: An adversarial attack $A_a^{i} \in A_{a}$ is a human-imperceptible noise ($\left\lVert \delta \right\rVert_{\infty} < \epsilon$) crafted to alter model’s predictions on the perturbed sample (i.e. adversarial sample) $x_{oa}$ from the original sample $x_o$. In black box adversarial attacks, the surrogate model $S_{m}$ creates adversarial samples, transferable to the black box model $B_{m}$. 

\textbf{Wavelet Transforms}: Wavelets represent the time series signal using linear combinations of an orthogonal basis depending on which there are different types of wavelets such as Haar, Cohen and Daubechies~\cite{daubechies}. The $2$D DWT on an $i^{th}$ image $x^i$ for level $1$ yields low pass subband (i.e approximation coefficients - denoting by $LL_{1}^i$) and high pass subband (i.e. detail-level coefficients - denoting by $LH_{1}^i$, $HL_{1}^i$, $HH_{1}^i$). In multi-level DWT, the approximation subband is further decomposed (for e.g. on a $2$-level decomposition, $LL_{1}^i$ is also decomposed to $LL_{2}^i$, $LH_{2}^i$, $HL_{2}^i$, $HH_{2}^i$).

\section{Proposed Approach}
\label{sec:proposed}
In this section, we first discuss the model stealing method (Sec.~\ref{subsec:model_and_data}) that we use to train the surrogate model $S_{m}$  (as proxy for black box model $B_{m}$) and generate synthetic data $S_{d}$ (as proxy for original training data $O_{d}^{train}$). Next, we propose our method to remove the detail coefficients (Sec.~\ref{subsec:WCSM}) that can be most corrupted by an adversarial attack and select the important coefficients to preserve the signal strength in terms of retaining feature discriminability. We dub this approach as wavelet noise remover (WNR) and the coefficients are selected using the wavelet coefficient selection module (WCSM). To boost the performance, we propose a U-Net-based regenerator network (Sec.~\ref{subsec:Regenerator_network}) that takes the output of WNR as input and is trained to output a regenerated image on which 
the surrogate model would yield similar features as the features of the clean sample. The different steps involved in our proposed method (\textit{DBMA}) for providing data-free adversarial defense in the black box settings are shown in Fig.~\ref{fig:approach}.

\subsection{Obtain Proxy Model and Synthetic Data}
\label{subsec:model_and_data}
Given a black box model $B_{m}$, our first step is to obtain a proxy model $S_{m}$ which can allow gradient backpropagation. To steal the functionality of $B_{m}$, $S_{m}$ can be trained using a model stealing technique. But we also do not have access to the original training samples $O_{d}^{train}$. Hence, we use a data-free model stealing technique~\cite{barbalau2020black} that trains a generator using proxy data to produce synthetic samples ($S_{d}$) on which the black box model $B_{m}$ gives high-confident predictions. The surrogate model $S_{m}$ is then trained on synthetic data $S_{d}$ under the guidance of $B_{m}$, where the model $S_{m}$ is enforced to mimic the predictions of model $B_{m}$. The trained $S_{m}$ and the generated data $S_{d}$ are used in next steps. 

\subsection{Noise Removal with Wavelet Coefficient Selection Module (WCSM)}
\label{subsec:WCSM}

For an $i^{th}$ sample of the composed synthetic data $S_{d}$ (i.e. $x_s^i$), its corresponding wavelet coefficients are obtained by $DWT$ operation on it. The approximate coefficients are the low frequencies that are least affected by the adversarial attack (shown in Fig.~\ref{fig:wavelet_bargraph} (B)). Thus, we retain these coefficients. For e.g. on level-$2$ discrete wavelet decomposition, $LL_{2}^i$ (approximate coefficients for $i^{th}$ sample) is kept. As the adversarial attack severely harms the detail coefficients, we determine the coefficients that can be most affected by it using WCSM for effective noise removal. 

Based on our observation that the least affected approximate coefficients often have high magnitude coefficients, indicating that the high magnitude detail coefficients can be a good measure for choosing which coefficients to retain. Thus, we arrange the detail coefficients based on magnitude (from high to low order) and retain the top-$k$ \% coefficients. The efficacy of choosing top-$k$ compared to different baselines (such as random-$k$ and bottom-$k$) is empirically verified in Sec.{\color{red}{1}}.  However, determining the suitable value of $k$ is a challenge and, if not properly chosen, can lead to a major bottleneck in clean/adversarial performance. To handle it, we propose a wavelet coefficient selection mechanism that carefully selects the value of $k$ so that decent performance can be obtained on both clean and adversarial data.  

We empirically estimate optimal $k$ using all the crafted synthetic training samples $S_{d}$ and their corresponding adversarial counterparts. We define a quantity label consistency rate ($LCR^k$) which is calculated for a particular value of $k$ using the following steps:
\begin{enumerate}
    \item Construct adversarial synthetic samples ($\{x_{sa}^{i}\}_{i=1}^{N}$) by using an adversarial attack $A_a^j \in A_{a}$ on the surrogate model $S_{m}$.
    \item Obtain approximate and detail coefficients for each adversarial synthetic sample using $DWT(x_{sa}^i, l), \forall i \in (1,\dots,N)$ where $l$ is the decomposition level.
    \item Craft spatial samples ($\bar{S}_{da}^{k} = \{\bar{x}_{sa}^{i}\}_{i=1}^{N}$) using $IDWT$ operation on complete approximate and selected top-$k$ detail coefficients corresponding to each adversarial synthetic sample (i.e. $S_{da} = x_{sa}^i, \forall i \in (1,\dots,N)$). 
For simplicity, we envelop the operations ($2$) and ($3$) and name them `wavelet noise remover' (WNR). In general, for a given input image and $k$ value, WNR applies $DWT$ on input where the approximate coefficients and the chosen top-$k$ \% detail coefficients are retained, whereas the non-selected detail coefficients are made to zeros. These coefficients are then passed to $IDWT$ to obtain the filtered spatial image. 
\item Perform WNR by repeating the steps $2$ and $3$ on clean samples $x_s$ to obtain $\bar{S}_{d}^{k}$.
    
    \item Compare the predictions of black box model $B_m$ on samples of $\bar{S}_{d}^{k}$ 
    and the corresponding samples in $S_{d}$. $LCR_C^k$ denotes the fraction of clean samples whose predictions match when top-$k$ \% coefficients are selected.

    
    \item Compare predictions of  black box model $B_m$ on samples of $\bar{S}_{da}^{k}$ 
    and the corresponding samples in $S_{d}$. $LCR_A^k$ denotes fraction of adversarial samples whose predictions match when top-$k$ \% coefficients are selected.

    \item Compute $LCR^k =    LCR_C^k + LCR_A^k$

    \item Calculate the rate of change of $LCR^k$ (i.e. $ROC^k$) as \\
    $ROC^k$ = $LCR^{k+1}$ - $LCR^k$
\end{enumerate}

\begin{figure}[htp]
\centering
\centerline{\includegraphics[width=0.45\textwidth]{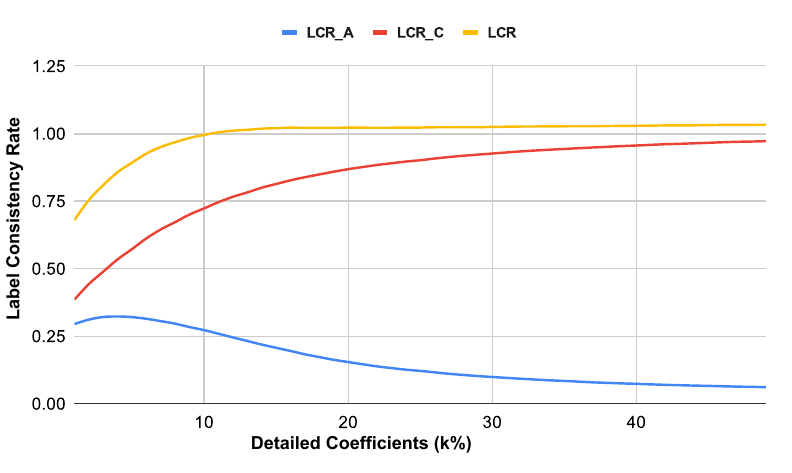}}
\vspace{-10pt}
\caption{Label consistency rates  ($LCR_A$ , $LCR_C$ and $LCR$)   vs detail coefficients ($k\%$) plotted using prediction from black-box model $B_{m}$ on Cifar-$10$ Dataset.
}
\label{fig: label_consistency_rate}
\end{figure}

\begin{figure}[htp]
\centering
\centerline{\includegraphics[width=0.45\textwidth]{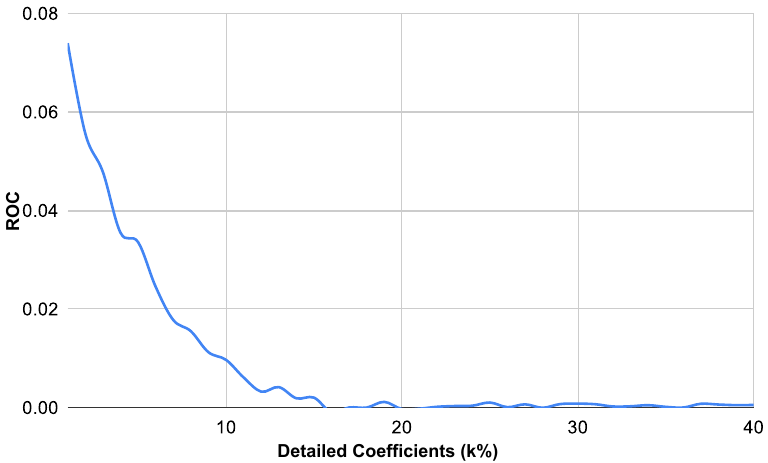}}
\vspace{-5pt}
\caption{Rate of change ($ROC$) of $LCR$ vs detail coefficients ($k\%$) plotted using prediction from black-box model $B_{m}$ on Cifar-$10$ data. As we increase value of $k$, $ROC$ becomes negligible. At $k=16$ it is close to zero. 
}
\label{fig: rate_of_change}
\end{figure}

Using above steps, we calculate $LCR$ for different values of $k \in [1, \cdots, k^{max}]$. As shown in Fig.~\ref{fig: label_consistency_rate}, we observe that as we increase the value of $k$, initially $LCR_A$ increases and reaches its maximum value, and then it starts decreasing. High $LCR_A$ implies that the predictions of the $B_m$ model on $\bar{S}_{da}^{k}$ and $S_{d}$ have a low mismatch. Similarly, with the increase in value of $k$, $LCR_C$ keeps increasing which implies as we add more coefficients, model discriminability increases. The value of $LCR$ increases with the value of $k$, but the rate of increase of $LCR$ keeps decreasing. Refer Fig.~\ref{fig: rate_of_change} where we plot the rate of change (ROC). We choose $\hat{k}$ at which ROC saturates. Here 
ROC is negligible at $k=16$. This value of $k$ gives the best trade-off between clean and adversarial accuracy (empirically validated in Sec.~\ref{subsec:coefficients_quantity}). Thus, wavelet noise remover (WNR) is applied on input at estimated optimal $k$ ($\hat{k}$).

\subsection{Training of Regenerator Network}
\label{subsec:Regenerator_network}
\vspace{-5pt}
After the noise removal using the WNR at optimal $k$ (i.e. $\hat{k}$ obtained by WCSM), there is also loss in the image signal information as a side effect. Thus, we further regenerate the coefficients using a regenerator network ($R_n$) that takes input as the output of WNR at $\hat{k}$ and yields reconstructed image which is finally passed to the black box model $B_{m}$ for test predictions. The architecture of our $R_n$ network is a U-net based architecture with skip connections which is inspired from~\cite{ronneberger2015u}. For more details on the architecture of regenerator network, refer Supplementary (Sec.{\color{red}{5}}).
\vspace{-5pt}
\begin{equation}
\small{\begin{gathered}
\hat{k}= WCSM(S_d, S_m)\\ \bar{x}_{s}^{i} = WNR(x_{s}^{i},  \hat{k}); \hspace{5pt} \bar{x}_{sa}^{i} = WNR(x_{sa}^{i},  \hat{k})\\
\end{gathered}}
\end{equation}
\vspace{-5pt}

For training, we feed the output obtained from the WNR at $\hat{k}$ as input to the network $R_n$ and obtain a reconstructed image which is passed to the frozen surrogate model $S_m$ for loss calculations. The losses used to train $R_n$ are as follows:\\
\noindent\textbf{Cosine similarity loss ($L_{cs}$)}: To enforce similar predictions from $S_m$ on the regenerated synthetic sample and the corresponding original synthetic sample. $L_{cs} = CS(S_m(R_n(\bar{x}_{s}^{i})),  S_m(x_{s}^{i}))$.\\
\noindent\textbf{KL divergence loss ($L_{kl}$)}: To align the predictions of $S_m$ on the regenerated sample and its adversarial counterpart $L_{kl} = KL(soft(S_m(R_n(\bar{x}_{sa}^{i}))), soft(S_m(R_n(\bar{x}_{s}^{i}))))$.\\ 
\noindent\textbf{Spatial consistency loss ($L_{sc}$)}: To make sure that the spatial reconstructed image (clean and adversarial) and the corresponding original synthetic image are similar in the image manifold $L_{sc} = \left\lVert R_n(\bar{x}_{s}^{i}) - x_{s}^{i} \right\rVert_{1} + \left\lVert R_n(\bar{x}_{sa}^{i}) - x_{s}^{i} \right\rVert_{1}$. 

Here, $CS$ and $KL$ denotes cosine similarity and KL divergence respectively. Overall loss used in training $R_n$:
\vspace{-5pt}
\begin{equation}
\begin{aligned}
L(R_n^{\theta}) = -\lambda_1 L_{cs} + \lambda_2 L_{kl} + \lambda_3 L_{sc}
\label{eq:total_loss}
\end{aligned}
\end{equation}
\vspace{-1pt}
Finally, the black box model $B_m$ is modified by prepending the WNR (with $k=\hat{k}$) and the trained $R_n$ network to it.  The resulting black box model defends the adversarial attacks which we discuss in detail in next section.

\section{Experiments}
\label{sec:experiments}
\vspace{-3pt}
In this section, we validate the effectiveness of our proposed method (DBMA) and perform ablations to show the importance of individual components. We use the benchmark classification datasets i.e. CIFAR-$10$~\cite{Krizhevsky2009LearningML} and SVHN~\cite{SVHN}, on which we evaluate the clean and the adversarial accuracy against three different adversarial attacks (i.e., BIM~\cite{kurakin2018adversarial}, ~PGD\cite{madry2017towards} and Auto Attack~\cite{croce2020reliable}). Unless it is mentioned, we use Alexnet~\cite{Alexnet} as black box $B_{m}$ (results on a larger black-box model are in Sec. {\color{red}{8}} in supplementary) and Resnet-$18$~\cite{he2016deep} as the defender’s surrogate model $S_{m}^d$, which the defender uses to train the regenerator network $R_n$ as explained in Sec.~\ref{sec:proposed}. In the black-box setting, attackers also do not have access to the black-box model’s weights, thus restricting the generation of adversarial samples. So similar to the defender, we leverage the model stealing techniques to get a new surrogate model $S_m^a$, which the attacker uses for generating the adversarial samples. While evaluating against different attacks, we assume the attacker has access to defense components, i.e., the attacker uses model stealing methods to steal the functionality of the defense components along with the victim model.

We perform experiments with two different architectures for $S_m^a$: Alexnet-half and Alexnet, which are similar to the black-box model (Alexnet), making it tough for the defender.  Ablation for different combination of $S_m^d$ and $S_{m}^a$ are in Sec .{\color{red}{7}} in supplementary. The attacker uses the same model stealing technique~\cite{barbalau2020black} as used by defender. It is important to note that our rigorous approach; we grant the attacker access to our exact model-stealing technique, architecture, and related details to ensure that any performance improvement is not attributed to differences in techniques or architecture between the attacker and defender. We use the Daubechies wavelet for both $DWT$ and $IDWT$ operations. Refer to supplementary (Sec. {\color{red}{2}}) for experimental results on other wavelets. The decomposition level is fixed at $2$ for all the experiments and ablations. The value of $k^{max}$ is taken as $50$. We assign equal weights to all the losses with weight as $1$ (i.e. $\lambda_1 = \lambda_2 = \lambda_3=1$) in eq.~\ref{eq:total_loss}. 
\vspace{-5pt}
\begin{figure}[htp]
\centering
\centerline{\includegraphics[width=0.4\textwidth]{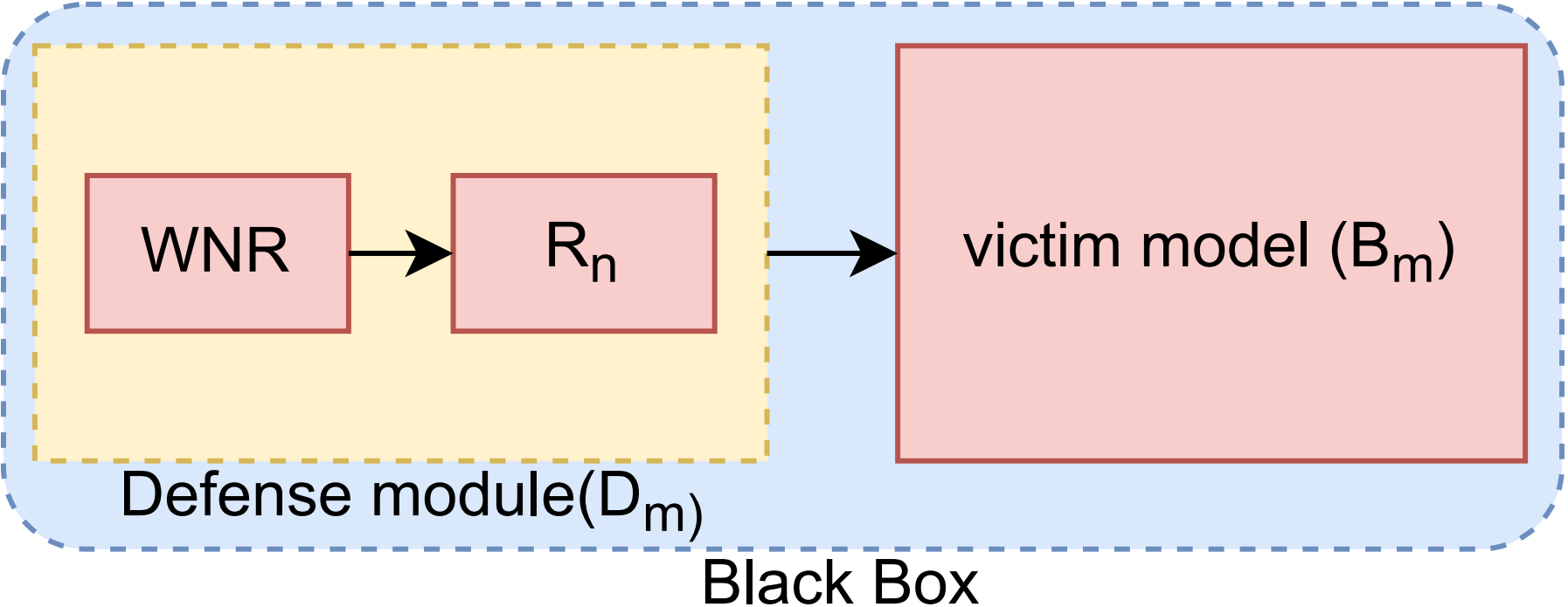}}
\vspace{-10pt}
\caption{Defense module $D_{m}$ consisting of Wavelet Noise Remover (WNR) and Regenerator $R_{n}$ is prepended before the victim model $B_{m}$ in our approach (DBMA). The $D_{m}$ and $B_{m}$ are combinedly considered as the black-box model by the attacker.}
\label{fig:Attacker_black_box}
\end{figure}

The defender constructs a defense module ($D_m$) using $S_m^d$. In subsections \ref{subsec:coefficients_quantity} and \ref{subsec:benefit_WNR}, the defense module only consists of the WNR , whereas subsections \ref{subsec:losses} onwards, both $R_{n}$ and WNR are part of the defense module as shown in Fig.~\ref{fig:Attacker_black_box}. We prepend the defense module before the $B_{m}$ to create a new black box model that is used to defend against the adversarial attacks. To show the efficacy of defense components used in our method (DBMA), we consider the most challenging scenario, where the attacker uses the same model stealing technique as defender, and considers the defense module also a part of the black-box model while generating adversarial samples. 
\subsection{Ablation on quantity of coefficients}
\vspace{-3pt}
\label{subsec:coefficients_quantity}
\vspace{-10pt}
\begin{table}[hbp]
\centering
\caption{\small{Investigating the efficacy of our proposed WCSM in determining the quantity of detail coefficients to retain. The case (No-$k$) yields poor results justifying the need to preserve detail coefficients. Unlike, low and high values of $k$, we obtain better trade-off between clean and adversarial performance on our-$k$.}}
\label{tab:table_2}
\scalebox{0.88}{
\begin{tabular}{|c|cccc|}
\hline
\multirow{-1.5}{*}{\begin{tabular}[c]{@{}c@{}}Amount of\\ detail \\ coefficients ($k$)\end{tabular}} &
  \multicolumn{4}{c|}{\begin{tabular}[c]{@{}c@{}}Black Box Model : Alexnet\\ Surrogate Model (defense): Resnet-18\end{tabular}} \\ \cline{2-5} 
                             & \multicolumn{1}{c|}{clean} & \multicolumn{1}{c|}{BIM}   & \multicolumn{1}{c|}{PGD}   & Auto Attack \\ \hline

 No $k$   & \multicolumn{1}{c|}{31.19} & \multicolumn{1}{c|}{5.88}  & \multicolumn{1}{c|}{4.68}  & 9.35        \\ 
                         low-$k$ ($k$=1) & \multicolumn{1}{c|}{42.75} & \multicolumn{1}{c|}{10.2}  & \multicolumn{1}{c|}{8.72}  & 15.8        \\ 
                         low-$k$ ($k$=2) & \multicolumn{1}{c|}{50.17} & \multicolumn{1}{c|}{15.37}  & \multicolumn{1}{c|}{14.14}  & 21.92        \\ 
                         low-$k$ ($k$=4) & \multicolumn{1}{c|}{59.14} & \multicolumn{1}{c|}{\textbf{17.54}}  & \multicolumn{1}{c|}{\textbf{16.03}}  & \textbf{25.08}        \\ 
                         high-$k$ ($k$=50) & \multicolumn{1}{c|}{82.58} & \multicolumn{1}{c|}{5.58}  & \multicolumn{1}{c|}{3.33}  & 10.44       \\ 
                         \textbf{our}-$k$ ($k$=16) & \multicolumn{1}{c|}{77.92} & \multicolumn{1}{c|}{{15.98}} & \multicolumn{1}{c|}{{14.04}} & {21.34}        \\ \hline
\end{tabular}
}

\end{table}

In Sec.~\ref{sec:intro} and \ref{sec:proposed}, we discussed the importance of selecting the optimal number of detail coefficients ($\hat{k}$) and proposed the steps to find the value of $\hat{k}$ using the WCSM module. In this subsection, we do an ablation over the different choices for values of $k$ (i.e., the number of detail coefficients to select) and analyze its impact on clean and adversarial accuracy. Specifically, we consider six distinct values of $k$ across a wide range i.e. $0$ (no detail coefficients, only approximate coefficients), $1, 2$ and $4$(small $k$), $50$(large $k$), $\hat{k}$ (optimal $k$ given by WCSM). Fig.~\ref{fig: rate_of_change} shows the graph of the rate of change of LCR for different values of $k$. We select the value of $k$ at which the 
ROC 
starts saturating, i.e. $k$ with value $16$ as $\hat{k}$. The results corresponding to the different values of $k$ are shown in Table~\ref{tab:table_2}. We observe poor performance for both adversarial and clean samples when no detail coefficients are taken. On increasing the fraction of detail coefficients ($k = 1,2,4$), an increasing trend for both clean and adversarial performance is observed. Further, for a high value of $k$ clean accuracy improves, but with a significant drop in the adversarial performance. Our choice of $k$ (i.e., $\hat{k}$) indeed leads to better clean accuracy with decent adversarial accuracy, hence justifying the importance of the proposed noise removal using WCSM component. 

\subsection{Effect of wavelet noise remover with WCSM}
\label{subsec:benefit_WNR}
In this section, we study the effect of prepending the WNR module to the black-box model $B_{m}$. WNR selects the approximate coefficients and optimal $\hat{k}\%$  detail coefficients (obtained by WCSM). It filters out the remaining detail coefficients, which helps in reducing the adversarial noise from the samples. We evaluate the performance of the $B_{m}$ with and without the WNR Module and present the results in Table~\ref{tab:table3}. When the attacker’s surrogate model is Alexnet-half, adversarial accuracy improves by $\approx19-22\%$ across attacks using the WNR module. Similarly, when the attacker’s surrogate model is Alexnet, the adversarial accuracy improves by $\approx11-12\%$. 

\begin{table}[htp]
\centering
\vspace{-5pt}
\caption{Our wavelet noise remover using WCSM yields improvement in adversarial accuracy with small drop in clean accuracy.}
\label{tab:table3}
\scalebox{0.85}{
\begin{tabular}{|c|c|cccc|}
\hline
\multirow{-1.5}{*}{\begin{tabular}[c]{@{}c@{}}Surrogate\\ model \\ (attacker)\end{tabular}} &
  \multirow{2}{*}{\begin{tabular}[c]{@{}c@{}}Noise Removal\\ using WCSM\end{tabular}} &
  \multicolumn{4}{c|}{\begin{tabular}[c]{@{}c@{}}Black Box Model : Alexnet\\ Surrogate Model (defense): Resnet-18\end{tabular}} \\ \cline{3-6} 
 &
   &
  \multicolumn{1}{c|}{clean} &
  \multicolumn{1}{c|}{BIM} &
  \multicolumn{1}{c|}{PGD} &
  Auto Attack \\ \hline
\multirow{2}{*}{\begin{tabular}[c]{@{}c@{}}Alexnet-\\ half\end{tabular}} &
 \multicolumn{1}{l|}{\hspace{0.53in} No}  &
  \multicolumn{1}{c|}{82.58} &
  \multicolumn{1}{c|}{7.02} &
  \multicolumn{1}{c|}{4.53} &
  \multicolumn{1}{c|}{11.65} \\ 
 &
  \multicolumn{1}{c|}{\hspace{0.05in} (\textbf{Ours}) Yes}  &
  \multicolumn{1}{c|}{77.92} & \multicolumn{1}{c|}{\textbf{26.66}} & \multicolumn{1}{c|}{\textbf{24.55}} & \textbf{34.02} \\ \hline
\multirow{2}{*}{Alexnet} &
 \multicolumn{1}{l|}{\hspace{0.53in} No} &
  \multicolumn{1}{c|}{82.58} &
  \multicolumn{1}{c|}{4.17} &
  \multicolumn{1}{c|}{2.19} &
  \multicolumn{1}{c|}{8.55} \\ 
 &
  \multicolumn{1}{c|}{\hspace{0.05in}(\textbf{Ours}) Yes} &
  \multicolumn{1}{c|}{77.92} & \multicolumn{1}{c|}{\textbf{15.98}} & \multicolumn{1}{c|}{\textbf{14.04}} & \textbf{21.34} \\ \hline
\end{tabular}
}

\end{table}
\vspace{-10pt}
\subsection{Ablation on losses}
\label{subsec:losses}
Until now, we performed experiments by using only the WNR defense module. Now, we additionally attach another defense module ($R_n$) to WNR (refer Fig~\ref{fig:Attacker_black_box}). In this subsection, we perform ablation to demonstrate the importance of different losses used for training the Regenerator network $R_{n}$. As shown in eq.~\ref{eq:total_loss}, the total loss $L$ is the weighted sum of three different losses (i.e., $L_{cs}$, $L_{kl}$, and $L_{sc}$). To determine the effect of each of the individual losses, we train $R_{n}$ using only the $L_{cs}$ loss,  $L_{kl}$ loss and $L_{sc}$ loss. Further to analyse the cumulative effect, we train $R_{n}$ with different possible pairs of loss i.e., $L_{cs}$ + $L_{sc}$ loss,  $L_{cs}$ + $L_{kl}$ loss,  $L_{kl}$ + $L_{sc}$ loss, and finally with the total loss ($L_{cs}$ + $L_{kl}$ + $L_{sc}$) respectively. The results are displayed in Table~\ref{tab:table-4}.
\begin{table}[hbp]
\centering
\caption{Contribution of different losses used for training $R_{n}$.
The loss ($L_{cs}$ + $L_{kl}$ + $L_{sc}$) gives the best improvement in adversarial accuracy with decent clean accuracy.}
\label{tab:table-4}
\scalebox{0.85}{
\begin{tabular}{|c|c|cccc|}
\hline
\multirow{-1.5}{*}{\begin{tabular}[c]{@{}c@{}}Surrogate \\ model \\ (attacker)\end{tabular}} &
  \multirow{-1.5}{*}{\begin{tabular}[c]{@{}c@{}}Losses to train\\ Regenerator \\ network ($R_{n}$)\end{tabular}} &
  \multicolumn{4}{c|}{\begin{tabular}[c]{@{}c@{}}Black Box Model : Alexnet\\ Surrogate Model (defense): Resnet-18\end{tabular}} \\ \cline{3-6} 
                         &                       & \multicolumn{1}{c|}{clean} & \multicolumn{1}{c|}{BIM}   & \multicolumn{1}{c|}{PGD}   & Auto Attack \\ \hline 
\multirow{3}{*}{\begin{tabular}[c]{@{}c@{}}Alexnet-\\ half\end{tabular}} &
  $L_{cs}$ &
  \multicolumn{1}{c|}{78.96} &
  \multicolumn{1}{c|}{26.33} &
  \multicolumn{1}{c|}{24.75} &
  33.81 \\ 
                         & $L_{sc}$          & \multicolumn{1}{c|}{78.85} & \multicolumn{1}{c|}{27.38} & \multicolumn{1}{c|}{25.75} & 35.51       \\ 
                         & $L_{kl}$          & \multicolumn{1}{c|}{9.82} & \multicolumn{1}{c|}{6.56} & \multicolumn{1}{c|}{6.54} & 8.98       \\ 
                         & $L_{cs}$ + $L_{sc}$ & \multicolumn{1}{c|}{79.75} & \multicolumn{1}{c|}{27.70} & \multicolumn{1}{c|}{25.34} & 34.64       \\ 
                         & $L_{cs}$ + $L_{kl}$ & \multicolumn{1}{c|}{62.06} & \multicolumn{1}{c|}{36.03} & \multicolumn{1}{c|}{35.93} & 43.05       \\ 
                         & $L_{kl}$ + $L_{sc}$ & \multicolumn{1}{c|}{65.94} & \multicolumn{1}{c|}{37.72} & \multicolumn{1}{c|}{37.62} & 46.14       \\ 
                         & $L_{cs}$+ $L_{kl}$ + $L_{sc}$ & \multicolumn{1}{c|}{73.77} & \multicolumn{1}{c|}{\textbf{42.71}} & \multicolumn{1}{c|}{\textbf{42.71}} & \textbf{50.63}       \\ \hline 
\multirow{3}{*}{Alexnet} & $L_{cs}$                 & \multicolumn{1}{c|}{78.96} & \multicolumn{1}{c|}{16.34} & \multicolumn{1}{c|}{14.57} & 21.81       \\ 
                         & $L_{sc}$          & \multicolumn{1}{c|}{78.85} & \multicolumn{1}{c|}{17.68} & \multicolumn{1}{c|}{15.97} & 23.77       \\ 
                         & $L_{kl}$          & \multicolumn{1}{c|}{9.82} & \multicolumn{1}{c|}{6.61} & \multicolumn{1}{c|}{6.38} & 8.98       \\ 
                         & $L_{cs}$+ $L_{sc}$          & \multicolumn{1}{c|}{79.75} & \multicolumn{1}{c|}{17.28} & \multicolumn{1}{c|}{15.6} & 23.59       \\ 
                         & $L_{cs}$+ $L_{kl}$          & \multicolumn{1}{c|}{62.06} & \multicolumn{1}{c|}{24.86} & \multicolumn{1}{c|}{25.91} & 32.26       \\ 
                         & $L_{kl}$+ $L_{sc}$          & \multicolumn{1}{c|}{65.94} & \multicolumn{1}{c|}{31.04} & \multicolumn{1}{c|}{31.05} & 38.26       \\ 
                         & $L_{cs}$+ $L_{kl}$ + $L_{sc}$ & \multicolumn{1}{c|}{73.77} & \multicolumn{1}{c|}{\textbf{33.31}} & \multicolumn{1}{c|}{\textbf{31.72}} & \textbf{40.56}       \\ \hline
\end{tabular}
}

\end{table}




\begin{table*}[!t]
\centering
\vspace{-15pt}
\caption{Utility of each component used in our method (DBMA)- Wavelet Noise Remover (WNR) and Regenerator Network $R_{n}$ on SVHN and CIFAR dataset. WNR with $R_n$ yields huge gains in adversarial performance compared to baseline and WNR alone.}
\label{tab:table-5}
\scalebox{0.8}{
\begin{tabular}{|c|c|l|llll|}
\hline
\multirow{2}{*}{\begin{tabular}[c]{@{}c@{}}Surrogate Model \\  (attacker)\end{tabular}} &
  \multirow{2}{*}{Dataset} &
  \multirow{2}{*}{Method} &
  \multicolumn{4}{c|}{\begin{tabular}[c]{@{}c@{}}Black Box Model : Alexnet\\ Surrogate Model (defender) : Resnet-18\end{tabular}} \\ \cline{4-7} 
 &
   &
   &
  \multicolumn{1}{c|}{clean} &
  \multicolumn{1}{c|}{BIM} &
  \multicolumn{1}{c|}{PGD} &
 \begin{tabular}[c]{@{}c@{}}Auto Attack\end{tabular} \\ \hline
\multirow{5}{*}{\begin{tabular}[c]{@{}c@{}}Alexnet-half\end{tabular}} &
  SVHN &
  \begin{tabular}[l]{@{}l@{}} Baseline\\ SIT\cite{guesmi2021sit}\\ RDG\cite{qiu2021efficient}\\ WNR (\textbf{Ours})\\ WNR+ $R_n$ (\textbf{Ours})\end{tabular} &
  \multicolumn{1}{l|}{\begin{tabular}[l]{@{}l@{}}94.49\\ 68.72\\ 92.68\\ 94.21\\ 90.91\end{tabular}} &
  \multicolumn{1}{l|}{\begin{tabular}[l]{@{}l@{}}44.26\\ 46.76 ({\color{blue}{$\uparrow$ 2.5}})\\ 55.01 ({\color{blue}{$\uparrow$ 10.75}})\\ 55.42 ({\color{blue}{$\uparrow$ 11.16}})\\ \textbf{68.63} ({\color{blue}{$\uparrow$ 24.37}})\end{tabular}} &
  \multicolumn{1}{l|}{\begin{tabular}[l]{@{}l@{}}44.21\\ 46.21 ({\color{blue}{$\uparrow$ 2}})\\ 54.62 ({\color{blue}{$\uparrow$ 10.41}})\\ 55.70 ({\color{blue}{$\uparrow$ 11.49}})\\ \textbf{68.60} ({\color{blue}{$\uparrow$ 24.39}})\end{tabular}} &
  \begin{tabular}[l]{@{}l@{}}46.79\\ 50.57 ({\color{blue}{$\uparrow$ 3.78 }})\\ 58.35 ({\color{blue}{$\uparrow$ 11.56}})\\ 58.47({\color{blue}{$\uparrow$ 11.68}})\\ \textbf{71.71} ({\color{blue}{$\uparrow$ 24.92}})\end{tabular} \\ \cline{2-7} 
 &
  CIFAR &
  \begin{tabular}[l]{@{}l@{}}Baseline\\ SIT\cite{guesmi2021sit}\\ RDG\cite{qiu2021efficient}\\ WNR (\textbf{Ours})\\ WNR+ $R_n$ (\textbf{Ours})\end{tabular} &
  \multicolumn{1}{l|}{\begin{tabular}[l]{@{}l@{}}82.58\\ 51.16\\ 67.58\\ 77.92\\ 73.77\end{tabular}} &
  \multicolumn{1}{l|}{\begin{tabular}[l]{@{}l@{}}7.02\\ 22.05 ({\color{blue}{$\uparrow$ 15.03}})\\ 19.99 ({\color{blue}{$\uparrow$ 12.97}})\\ 26.66 ({\color{blue}{$\uparrow$ 19.64}})\\ \textbf{42.71} ({\color{blue}{$\uparrow$ 35.69}})\end{tabular}} &
  \multicolumn{1}{l|}{\begin{tabular}[l]{@{}l@{}}4.53\\ 21.68 ({\color{blue}{$\uparrow$ 17.15}})\\ 18.97 ({\color{blue}{$\uparrow$ 14.44}})\\ 24.55 ({\color{blue}{$\uparrow$ 20.02}})\\ \textbf{42.71} ({\color{blue}{$\uparrow$ 38.18}}) \end{tabular}} &
  \begin{tabular}[l]{@{}l@{}}11.65\\ 29.08 ({\color{blue}{$\uparrow$ 17.43}})\\ 29.64 ({\color{blue}{$\uparrow$ 17.99}})\\ 34.02 ({\color{blue}{$\uparrow$ 22.37}})\\ \textbf{50.63} ({\color{blue}{$\uparrow$ 38.98}})\end{tabular} \\ \hline
\multirow{5}{*}{Alexnet} &
  SVHN &
  \begin{tabular}[l]{@{}l@{}}Baseline\\ SIT\cite{guesmi2021sit}\\ RDG\cite{qiu2021efficient}\\ WNR (\textbf{Ours})\\ WNR+ $R_n$ (\textbf{Ours})\end{tabular} &
  \multicolumn{1}{l|}{\begin{tabular}[l]{@{}l@{}}94.49\\ 68.72\\ 92.68\\ 94.21\\ 90.91\end{tabular}} &
  \multicolumn{1}{l|}{\begin{tabular}[l]{@{}l@{}}38.14\\ 43.48 ({\color{blue}{$\uparrow$ 5.34}})\\ 50.28 ({\color{blue}{$\uparrow$ 12.14}})\\ 48.98 ({\color{blue}{$\uparrow$ 10.84}})\\ \textbf{63.13} ({\color{blue}{$\uparrow$ 24.99}})\end{tabular}} &
  \multicolumn{1}{l|}{\begin{tabular}[l]{@{}l@{}}38.19\\ 43.77 ({\color{blue}{$\uparrow$ 5.58}})\\ 50.29 ({\color{blue}{$\uparrow$ 12.1}})\\ 49.02 ({\color{blue}{$\uparrow$ 10.83}})\\ \textbf{63.12} ({\color{blue}{$\uparrow$ 24.93}})\end{tabular}} &
  \multicolumn{1}{l|}{\begin{tabular}[l]{@{}l@{}}40.16\\ 47.63 ({\color{blue}{$\uparrow$ 7.47}})\\ 53.79 ({\color{blue}{$\uparrow$ 13.63}})\\ 51.49 ({\color{blue}{$\uparrow$ 11.33}})\\ \textbf{66.18} ({\color{blue}{$\uparrow$ 26.02}})\end{tabular}} \\ 
  \cline{2-7} 
 &
  CIFAR &
  \begin{tabular}[l]{@{}l@{}}Baseline\\ SIT\cite{guesmi2021sit}\\ RDG\cite{qiu2021efficient}\\ WNR (\textbf{Ours})\\ WNR+ $R_n$ (\textbf{Ours})\end{tabular} &
  \multicolumn{1}{l|}{\begin{tabular}[l]{@{}l@{}}82.58\\ 51.16\\ 67.58\\ 77.92\\ 73.77\end{tabular}} &
  \multicolumn{1}{l|}{\begin{tabular}[l]{@{}l@{}}4.17\\ 18.64 ({\color{blue}{$\uparrow$ 14.47}})\\ 15.54 ({\color{blue}{$\uparrow$ 11.37}})\\ 15.98 ({\color{blue}{$\uparrow$ 11.81}})\\ \textbf{33.31} ({\color{blue}{$\uparrow$ 29.14}})\end{tabular}} &
  \multicolumn{1}{l|}{\begin{tabular}[l]{@{}l@{}}2.19\\ 18.43 ({\color{blue}{$\uparrow$ 16.24}})\\ 14.50 ({\color{blue}{$\uparrow$ 12.31}})\\ 14.04 ({\color{blue}{$\uparrow$ 11.85}})\\ \textbf{31.72} ({\color{blue}{$\uparrow$ 29.53}})\end{tabular}} &
  \begin{tabular}[l]{@{}l@{}}8.55\\ 26.23 ({\color{blue}{$\uparrow$ 17.68}})\\ 24.01 ({\color{blue}{$\uparrow$ 15.46}})\\ 21.34 ({\color{blue}{$\uparrow$ 12.79}})\\ \textbf{40.56} ({\color{blue}{$\uparrow$ 32.01}})\end{tabular} \\ \hline
\end{tabular}
}

\end{table*}
\vspace{-10pt}
Compared to the earlier best performance with WNR defense module (Table~\ref{tab:table3}), we observe 
$R_{n}$ trained with only $L_{cs}$ loss gives no significant improvement in both the adversarial and clean accuracy. Similar trend is observed for $R_{n}$ trained with $L_{sc}$ loss. However, using only the $L_{kl}$ loss shows a deteriorated clean and adversarial performance of $R_{n}$. Further, using the combination of both $L_{cs}$ and $L_{sc}$ loss also does not show much improvement. Combining the $L_{kl}$ with $L_{cs}$ and $L_{sc}$ loss improves the adversarial performance of $R_{n}$ appreciably, but with a drop in the clean accuracy. 
$L_{kl}$ with $L_{cs}$ loss shows a consistent improvement of $\approx9-11\%$ in adversarial accuracy across attacks using both the Alexnet and Alexnet-half. Similarly, the combination of $L_{kl}$ with $L_{sc}$ loss improves the adversarial accuracy of $R_{n}$ by $\approx11-13\%$ and $\approx15-17\%$ against the attacks using Alexnet-half and Alexnet respectively. When $R_{n}$ is trained with all three losses gives the best adversarial accuracy across all the possible combinations. We observe an overall improvement of $\approx16-19\%$ across different attacks using both Alexnet and Alexnet-half with a slight drop in clean accuracy ($\approx4\%$).  Regenerator network regenerates the lost coefficients, but as explained in section~\ref{subsec:WCSM}, detail coefficients also cause a decrease in adversarial accuracy. When we train a regenerator network using the combination of $L_{cs}$, $L_{sc}$, and $L_{kl}$ loss, the  $L_{cs}$ and $L_{sc}$  loss help to increase clean accuracy, but at the same time $L_{kl}$ loss ensures regenerated coefficients do not decrease the adversarial accuracy. To achieve best tradeoff between clean and adversarial accuracy, $R_n$ gets trained to increase adversarial accuracy at the cost of decreased clean accuracy compared to $R_n$ network trained with only $L_{cs}$ loss. 

\subsection{Comparison with existing Data and Training efficient defense methods}
\label{subsec:datasets}
In this subsection, we validate the efficacy of our proposed method DBMA by comparing it with two other state-of-the-art defense methods: SIT\cite{guesmi2021sit} and GD\cite{qiu2021efficient}. For comparison, we do experiments on two benchmark datasets, i.e., SVHN and CIFAR-$10$. We obtain the optimal $\hat{k}$ as $20$ using the WCSM module for the black box model trained on the SVHN dataset. In Table~\ref{tab:table-5}, while defending with SIT, the adversarial accuracy improves by $\approx2-3\%$ and $\approx5-7\%$ across attacks crafted using different $S_m^a$ (Alexnet and Alexnet-half) with a corresponding drop of $\approx2-3\%$ in clean accuracy. On the other hand, defending with GD, improves adversarial accuracy by $\approx2-3\%$ across attacks with a marginal drop of 1.66\% in clean accuracy. While defending with only WNR in the defender module, the adversarial accuracy improves by $\approx10-11\%$ across attacks crafted using different surrogate architectures $S_m^a$ (Alexnet and Alexnet-half). The clean accuracy, however, experiences a minor drop of less than $1\%$. By utilizing both the WNR and $R_{n}$ in the defender module, the adversarial performance further improves by $\approx13-14\%$ across the attacks with a drop of $\approx4\%$ in clean accuracy. Overall, we observe a gain of $\approx24-26\%$ in adversarial accuracy compared to the baseline model, at the cost $\approx5\%$ drop in clean accuracy.
Similarly, for the CIFAR-10 dataset, we observe an overall improvement of  $\approx35-38\%$ and $\approx29-32\%$ against attacks crafted using Alexnet-half and Alexnet, respectively. However, defending with SIT and RDG only improves the adversarial accuracy by  $\approx12-17\%$. Additionally, the clean accuracy drops by $\approx31\%$ and $\approx15\%$ when using SIT and RDG, respectively. In comparison, when using DBMA, we observed a relatively small drop of $\approx8\%$ in clean accuracy, which is reasonable considering the challenging nature of our problem setup. Even in traditional adversarial training with access to full data, clean performance often drops at the cost of improving adversarial accuracy~\cite{madry2017towards}. In our case, neither the training data nor the model weights are provided. Moreover, the black-box model is often obtained as APIs, and re-training the model from scratch becomes unfeasible. Considering these difficulties, the drop we observe on clean data is small with respectable overall performance.
\vspace{-5pt}

\section{Conclusion}
\vspace{-2pt}
We introduced DBMA, a novel defense strategy to defend black-box models from adversarial attacks without relying on training data. DBMA incorporates two defense components: a) Wavelet Noise Remover (WNR) that removes the most contaminated areas by adversarial attacks while preserving less affected regions b) a Regenerator network to restore lost information post WNR noise removal. Through various ablations and experiments, we demonstrated efficacy of each defense component. Our method DBMA significantly enhances robustness against data-free black box attacks across datasets and against diverse model-stealing methods. Similar to adversarial defenses in white-box setups, we observe that significant gains in robust accuracy come at the cost of a slight drop in clean accuracy. In future work, we plan to work on further mitigating this trade-off.

{
    \bibliographystyle{ieeenat_fullname}
    \bibliography{main}
}

\clearpage
\newpage
\setcounter{section}{0}
\setcounter{table}{0}
\setcounter{figure}{0}
\setcounter{equation}{0}
\begin{center}
    \LARGE{\textbf{{\textit{Supplementary for:} \\``Data-free Defense of Black Box Models Against Adversarial Attacks"}}}

\end{center}

\vspace{18pt}
\hrule
\vspace{18pt}

\section{Ablation on coefficient selection strategy}
\label{subsec:coefficient_selection}
In our method DBMA, we select only a limited number of coefficients(i.e., $k\%$ coefficients) to obtain a decent tradeoff between the adversarial and clean performance. As shown in Fig~1 (A) in the main paper, the least affected approximate coefficients have a high magnitude. Thus, we select the essential detail coefficients as the top-$k$ high magnitude. As they are likely to be less affected by the adversarial attack (Fig~1 (B)) (main paper), they can yield better performance. In this section, we perform an ablation to assess the effectiveness of selecting top-$k$ coefficients over the other possible choices. For comparative analysis, we perform the experiments using the bottom-$k$ and random-$k$ coefficients. Table~\ref{tab:table_1} shows the adversarial and clean performance for the different coefficient selection strategies. The top-$k$ coefficient selection strategy, selecting the most important coefficient in terms of magnitude, improves both the clean and adversarial performance. On the other hand bottom-$k$ coefficient selection strategy, selecting the least significant coefficients shows the least performance as they primarily consist of contaminated high-frequency content. Although, randomly selecting $k\%$ coefficients showed improved performance than the bottom-$k$, it still performs poorly compared to top-$k$ coefficient selection strategy.

\begin{table}[htp]
\centering
\caption{Performance comparison when $k\%$ detail coefficients are selected in wavelet coefficient selection module (WCSM) using different methods. Selection of top-k coefficients yields better clean and adversarial accuracy than other strategies.}
\label{tab:table_1}
\scalebox{0.8}{
\begin{tabular}{|c|c|cccc|}
\hline
  \multirow{-1.5}{*}{\begin{tabular}[c]{@{}c@{}}Surrogate\\ model \\ (attacker)\end{tabular}} &
  \multirow{-1.5}{*}{\begin{tabular}[c]{@{}c@{}}coefficient\\ selection\\ strategy\end{tabular}}  & \multicolumn{4}{c|}{\begin{tabular}[c]{@{}c@{}}Black Box Model : Alexnet\\ Surrogate Model (defense): Resnet-18\end{tabular}} \\ \cline{3-6} 
  & &
  \multicolumn{1}{c|}{clean} &
  \multicolumn{1}{c|}{BIM} &
  \multicolumn{1}{c|}{PGD} &
  Auto Attack \\ \hline
   & bottom-$k$ & \multicolumn{1}{c|}{31.25}          & \multicolumn{1}{c|}{8.59}          & \multicolumn{1}{c|}{7.32}           & 12.57         \\ 
   & random-$k$ & \multicolumn{1}{c|}{42.58}          & \multicolumn{1}{c|}{13.13}         & \multicolumn{1}{c|}{11.74}          & 19.77         \\ 
\multirow{-3}{*}{{\begin{tabular}[c]{@{}c@{}}Alexnet-\\ half\end{tabular}}} &
  top-$k$ (\textbf{ours}) &
  \multicolumn{1}{c|}{\textbf{77.92}} &
  \multicolumn{1}{c|}{\textbf{26.66}} &
  \multicolumn{1}{c|}{\textbf{24.55}} &
  \textbf{34.02} \\ \hline
                          & bottom-$k$ & \multicolumn{1}{c|}{31.25}          & \multicolumn{1}{c|}{6.14}          & \multicolumn{1}{c|}{4.84}           & 10.92          \\ 
                          & random-$k$ & \multicolumn{1}{c|}{42.92}          & \multicolumn{1}{c|}{8.61}          & \multicolumn{1}{c|}{7.52}          & 14.29         \\ 
\multirow{-3}{*}{Alexnet} & top-$k$ (\textbf{ours})    & \multicolumn{1}{c|}{\textbf{77.92}}          & \multicolumn{1}{c|}{\textbf{15.98}}         & \multicolumn{1}{c|}{\textbf{14.04}}         & \textbf{21.34}          \\ \hline
\end{tabular}
}

\end{table}

\section{Performance of our method (DBMA) using different wavelets}
\label{supp_sec:diff_wavs}

The experiments in the main draft, have used Daubechies wavelets~\cite{Daubechies1992TenLO} in wavelet noise remover (WNR). Along with Daubechies, several other wavelets are available in the literature that varies in time, frequency, and rate of decay. This section analyzes the effect of different wavelets on the performance of proposed data-free black box defense (DBMA). We perform experiments with Coiflets~\cite{Beylkin1991FastWT} and Biorthogonal wavelets~\cite{Cohen1992BiorthogonalBO}.  Table~\ref{tab:table1}  summarizes the results obtained with Resnet18 as the defender's surrogate model and Alexnet-half as the attacker's surrogate model. We observe a similar performance trend over the adversarial and clean samples on using the different wavelet functions when compared to the Daubechies. This confirms DBMA yields consistent performance irrespective of the choice of the wavelet function used.

\begin{table}[htp]
\centering
\normalsize
\caption{Ablation over different choices of wavelets that are used in wavelet noise remover (WNR). The performance of DBMA remains consistent across different choices of wavelet functions.}
\label{tab:table1}
\scalebox{0.8}{
\begin{tabular}{|c|c|cccc|}
\hline
\multirow{-1.5}{*}{\begin{tabular}[c]{@{}c@{}}Surrogate\\ model \\ (Attacker)\end{tabular}} &
  \multirow{-1.5}{*}{\begin{tabular}[c]{@{}c@{}}Surrogate\\ model \\ (Defender)\end{tabular}} &
  \multicolumn{4}{c|}{\begin{tabular}[c]{@{}c@{}}Black Box Model : Alexnet\\ Surrogate Model (defender): Resnet-18\end{tabular}} \\ \cline{3-6} 
 &            & \multicolumn{1}{c|}{clean} & \multicolumn{1}{c|}{BIM} & \multicolumn{1}{c|}{PGD} & Auto Attack \\ \hline
\multirow{3}{*}{\begin{tabular}[c]{@{}c@{}}Alexnet-\\ half\end{tabular}} &
  Biorthogonal &
  \multicolumn{1}{c|}{73.27} &
  \multicolumn{1}{c|}{42.51} &
  \multicolumn{1}{c|}{41.72} &
  \multicolumn{1}{c|}{51.11} 
   \\ 
 & Daubechies & \multicolumn{1}{c|}{73.77}      & \multicolumn{1}{c|}{42.71}    & \multicolumn{1}{c|}{42.71}    &
 \multicolumn{1}{c|}{50.63} \\ 
 & Coiflets   & \multicolumn{1}{c|}{73.14}      & \multicolumn{1}{c|}{42.51}    & \multicolumn{1}{c|}{44.22}    &
 \multicolumn{1}{c|}{51.94}
 \\ \hline
\end{tabular}
}

\end{table}
\vspace{-0.15in}
\section{Defense against different data-free black box attacks}
\label{subsec:data_free_black_box}
\begin{figure}[t]
\centering
\centerline{\includegraphics[width=0.45\textwidth]{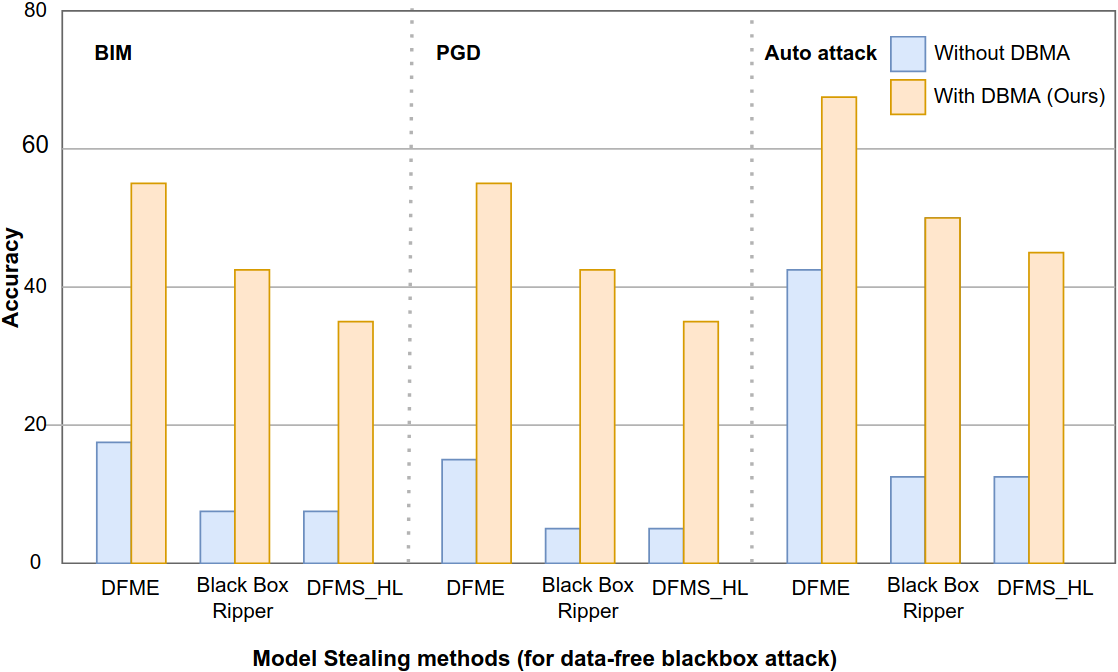}}
\caption{Performance of our approach DBMA for different model stealing methods used to get the attacker’s surrogate model for data-free black box attacks. DBMA consistently improves performance against different attacks 
across all model stealing methods.
}
\label{fig:different_model_stealing}
\end{figure}

In all the previous experiments we assume that both defender and attacker use the same model stealing technique~\cite{barbalau2020black} for creating a surrogate model. To prove our data-free black box defense (DBMA) is robust to different model stealing strategies, we evaluate our method against two different approaches: a) Data-free model extraction (DFME)~\cite{truong2021data} and b) Data-free model stealing in hard label setting (DFMS-HL)~\cite{sanyal2022towards} that are used by attacker to obtain a surrogate model (Alexnet-Half) for crafting adversarial samples. 
 As shown in Fig.~\ref{fig:different_model_stealing}, our method yields a consistent boost in adversarial accuracy on different data-free model stealing methods. In the case of DFME, we observe a massive improvement of $\approx$$36\%$, $\approx$$40\%$ and $\approx$$21\%$ on BIM , PGD and Auto Attack respectively. On the other hand, in case of DFMS-HL the performance against the three attacks improves by $\approx$$28\%$, $\approx$$29\%$, and $\approx$$31\%$. 
Overall across different model stealing methods, our method (DBMA) yields significant improvement in adversarial accuracy (i.e. $\approx 29-42\%$ in PGD, $\approx 28-38\%$ in BIM , and $\approx 25-31\%$ in state-of-the-art Auto Attack). 

\begin{table*}[htp]
\centering
\caption{Performance of DBMA across several data-free black box attacks that are constructed using the surrogate model having same architecture as the black box model. We observe consistent significant improvements in adversarial performance using proposed DBMA across different adversarial attacks and model stealing techniques.}
\label{tab:table2}
\scalebox{0.8}{
\begin{tabular}{|c|c|c|cccc|}
\hline
\multirow{-1.5}{*}{\begin{tabular}[c]{@{}c@{}}Surrogate \\ model \\ (attacker)\end{tabular}} &
  \multirow{2}{*}{Method} &
  \multirow{2}{*}{\begin{tabular}[c]{@{}c@{}}Model stealing \\ (Attacker)\end{tabular}} &
  \multicolumn{4}{c|}{\begin{tabular}[c]{@{}c@{}}Black Box Model : Alexnet\\ Surrogate Model (defense): Resnet-18\end{tabular}} \\ \cline{4-7} 
                         &                  &                  & \multicolumn{1}{c|}{clean} & \multicolumn{1}{c|}{BIM}   & \multicolumn{1}{c|}{PGD}   & Auto Attack \\ \hline
\multirow{6}{*}{Alexnet} & \multicolumn{1}{l|}{Without DBMA}     & Black Box Ripper & \multicolumn{1}{l|}{82.58} & \multicolumn{1}{l|}{4.17}  & \multicolumn{1}{l|}{2.19}  & \multicolumn{1}{l|}{8.55}        \\ 
                         & \multicolumn{1}{l|}{With DBMA (\textbf{Ours})} & Black Box Ripper & \multicolumn{1}{c|}{73.77} & \multicolumn{1}{l|}{\textbf{33.31} ({\color{blue}{$\uparrow$ 29.14}})} & \multicolumn{1}{l|}{\textbf{31.72} ({\color{blue}{$\uparrow$ 29.53}})} & 
                         \multicolumn{1}{l|}{\textbf{40.56} ({\color{blue}{$\uparrow$ 32.01}})}       \\ \cline{2-7} 
                         & \multicolumn{1}{l|}{Without DBMA}     & DFME             & \multicolumn{1}{c|}{82.58} & \multicolumn{1}{l|}{4.21}  & \multicolumn{1}{l|}{1.99}  & \multicolumn{1}{l|}{16.93}       \\ 
                         & With DBMA (\textbf{Ours}) & DFME             & \multicolumn{1}{c|}{73.77} & \multicolumn{1}{l|}{\textbf{42.84} ({\color{blue}{$\uparrow$ 38.63}})} & \multicolumn{1}{l|}{\textbf{42.49} ({\color{blue}{$\uparrow$ 40.5}})} & \multicolumn{1}{l|}{\textbf{55.04} ({\color{blue}{$\uparrow$ 38.11}})}       \\ \cline{2-7} 
                         & \multicolumn{1}{l|}{Without DBMA}     & DFMS-HL          & \multicolumn{1}{l|}{82.58} & \multicolumn{1}{l|}{3.30} & \multicolumn{1}{l|}{1.82} & \multicolumn{1}{l|}{7.84}       \\ 
                         & With DBMA (\textbf{Ours}) & DFMS-HL          & \multicolumn{1}{c|}{73.77} & \multicolumn{1}{l|}{\textbf{26.0} ({\color{blue}{$\uparrow$ 22.7}})} & \multicolumn{1}{l|}{\textbf{24.94} ({\color{blue}{$\uparrow$ 23.12}})} & \multicolumn{1}{l|}{\textbf{32.57} ({\color{blue}{$\uparrow$ 24.73}})}        \\ \hline
\end{tabular}
}

\end{table*}

Further, we check our defense in a more tougher scenario, where attacker is aware of the black box model $B_m$'s architecture (i.e., Alexnet), and uses the same for attacker's surrogate model ($S_m^a$). The results for this setup are reported in Table~\ref{tab:table2}. Compared to baseline, we observe an improvement of $\approx29-32\%$, $\approx38-40\%$ and $\approx22-25\%$ in adversarial accuracy across attacks using the Black box ripper, DFME and DFMS-HL methods, respectively. This indicates that even if the attacker is aware of the black box model's architecture, our method DBMA can provide data-free black-box adversarial defense irrespective of the different model stealing methods.

Hence, 
our proposed method DBMA provides strong robustness even when the attacker uses a different model stealing strategy compared to the defender. Refer to next section for adversarial robustness results when the defender uses different model stealing methods to obtain the surrogate model.

\section{DBMA using different model stealing methods} 
In all the experiments in the main draft, we used the Black Box Ripper (BBR) model stealing method to obtain surrogate model for defense. To demonstrate that our method DBMA can work across different model stealing techniques also, we 
obtain the defender's surrogate model using a different model stealing method (DFMS-HL). To get better insights, we analyze the performance of DBMA by varying the model stealing techniques for both attack and defense. 

In Table \ref{tab:table_3} we observe the clean accuracy is not affected on using different model stealing techniques. DBMA obtains the best performance when the attacker uses BBR to attack while the defense is performed using model stealing DFMS-HL (third row). However, when the attacker uses the DFMS-HL method to create a surrogate model (second and fourth row), the adversarial accuracy decreases by $\approx 7-11\%$ across attacks compared to the performance with BBR method used by attacker. For attacks crafted using BBR, the defender's performance with DFMS-HL remains almost similar to BBR (first and third rows). These results suggest that the adversarial samples created using the DFMS-HL are stronger than the ones created using BBR method. This aligns with the Sec. $5.6$ in the main draft, where we observed a similar trend indicating that the attacks crafted using DFMS-HL are powerful than BBR. Further, 
we observe that the choice of the defender’s model stealing method does not significantly affect the adversarial and clean accuracy.

Overall, for different model stealing methods, DBMA ensures good clean performance with decent adversarial performance.

\begin{table*}[h]
\centering
\caption{Performance of DBMA across different model stealing methods used by defender and attacker. DBMA obtains respectable performance irrespective of the model stealing technique used by either the defender or attacker.}

\label{tab:table_3}
\scalebox{0.9}{
\begin{tabular}{|l|l|l|llll|}
\hline
\multirow{-1.5}{*}{\begin{tabular}[c]{@{}c@{}}Surrogate\\ model \\ (attacker)\end{tabular}} &
  \multirow{-1.5}{*}{\begin{tabular}[c]{@{}c@{}}Model \\ stealing \\ (defender)\end{tabular}} &
  \multirow{-1.5}{*}{\begin{tabular}[c]{@{}c@{}}Model \\ stealing \\ (attacker)\end{tabular}} &
  \multicolumn{4}{c|}{\begin{tabular}[c]{@{}c@{}}Black Box Model : Alexnet\\ Surrogate Model (defense): Resnet-18\end{tabular}} \\ \cline{4-7} 
                              &                  &                  & \multicolumn{1}{c|}{clean} & \multicolumn{1}{c|}{BIM}   & \multicolumn{1}{c|}{PGD}   & Auto Attack \\ \hline
\multirow{4}{*}{\begin{tabular}[c]{@{}c@{}}Alexnet-\\ half\end{tabular}} & BBR & BBR & \multicolumn{1}{c|}{73.77} & \multicolumn{1}{c|}{{42.71}} & \multicolumn{1}{c|}{{42.71}} & {50.63}       \\ 
                              & BBR & DFMS-HL          & \multicolumn{1}{c|}{73.77} & \multicolumn{1}{c|}{35.4} & \multicolumn{1}{c|}{34.58} & 43.05       \\ \cline{2-7} 
                              & DFMS-HL          & BBR & \multicolumn{1}{c|}{72.45} & \multicolumn{1}{c|}{\textbf{42.75}} & \multicolumn{1}{c|}{\textbf{43.08}} & \textbf{51.0}       \\ 
                              & DFMS-HL          & DFMS-HL          & \multicolumn{1}{c|}{72.45} & \multicolumn{1}{c|}{32.85} & \multicolumn{1}{c|}{31.86} & 40.6       \\ \hline
\end{tabular}
}

\end{table*}
\newcommand\myeq{\mkern1.5mu{=}\mkern1.5mu}

\begin{table*}[htp]
\centering
\caption{Performance of DBMA with and without regenerator network across different values of $k$. For low values, regenerator network improves both clean and adversarial accuracy. For $k\myeq16$,  small decrease in clean accuracy, but adversarial accuracy increases significantly.}
\label{tab:rn_network}
\scalebox{0.9}{
\begin{tabular}{|c|c|c|cccc|}
\hline
\multirow{-1.5}{*}{\begin{tabular}[c]{@{}c@{}}Surrogate \\ model \\ (attacker)\end{tabular}} &
  \multirow{2}{*}{\begin{tabular}[c]{@{}c@{}}Coefficients 
  \\($k\%$)\end{tabular}} &
  \multirow{2}{*}{\begin{tabular}[c]{@{}c@{}}Method\end{tabular}} &
  \multicolumn{4}{c|}{\begin{tabular}[c]{@{}c@{}}Black Box Model : Alexnet\\ Surrogate Model (defense): Resnet-18\end{tabular}} \\ \cline{4-7} 
                         &                  &                  & \multicolumn{1}{c|}{clean} & \multicolumn{1}{c|}{BIM}   & \multicolumn{1}{c|}{PGD}   & Auto Attack \\ \hline
\multirow{12}{*}{Alexnet-half} & \multirow{2}{*}{1}     & \multicolumn{1}{l|}{WNR} & \multicolumn{1}{l|}{42.75} & \multicolumn{1}{l|}{14.89}  & \multicolumn{1}{l|}{13.79}  & 21.41        \\ 
                         &  & WNR + $R_n$ & \multicolumn{1}{c|}{56.08} & \multicolumn{1}{c|}{30.58 ({\color{blue}{$\uparrow$ 15.69}})} & \multicolumn{1}{c|}{30.14 ({\color{blue}{$\uparrow$ 16.35}})} & 37.04 ({\color{blue}{$\uparrow$ 15.63}})   \\ \cline{2-7} 
                         & \multirow{2}{*}{2}  & \multicolumn{1}{l|}{WNR}             & \multicolumn{1}{l|}{50.17} & \multicolumn{1}{l|}{17.34}  & \multicolumn{1}{l|}{16.38}  & \multicolumn{1}{l|}{25.36}       \\ 
                     &  & WNR + $R_n$ & \multicolumn{1}{c|}{60.35} & \multicolumn{1}{c|}{34.94 ({\color{blue}{$\uparrow$ 17.60}})} & \multicolumn{1}{c|}{34.61 ({\color{blue}{$\uparrow$ 18.23}})} & 41.61 ({\color{blue}{$\uparrow$ 16.25}})       \\ \cline{2-7} 
                         & \multirow{2}{*}{4}  & \multicolumn{1}{l|}{WNR}             & \multicolumn{1}{c|}{59.14} & \multicolumn{1}{l|}{21.99}  & \multicolumn{1}{l|}{20.77}  & \multicolumn{1}{l|}{29.43}       \\ 
                         &  & WNR + $R_n$ & \multicolumn{1}{c|}{65.82} & \multicolumn{1}{c|}{39.65 ({\color{blue}{$\uparrow$ 17.66}})} & \multicolumn{1}{c|}{39.62 ({\color{blue}{$\uparrow$ 18.85}})} & 46.94 ({\color{blue}{$\uparrow$ 17.51}})       \\ \cline{2-7} 
                         
                        & \multirow{2}{*}{8}  & \multicolumn{1}{l|}{WNR}             & \multicolumn{1}{c|}{69.89} & \multicolumn{1}{l|}{24.9}  & \multicolumn{1}{l|}{23.54}  & \multicolumn{1}{l|}{33.04}       \\ 
                         &  & WNR + $R_n$ & \multicolumn{1}{c|}{70.37} & \multicolumn{1}{c|}{41.89 ({\color{blue}{$\uparrow$ 16.99}})} & \multicolumn{1}{c|}{42.21 ({\color{blue}{$\uparrow$ 18.67}})} & 49.88 ({\color{blue}{$\uparrow$ 16.84}})      \\ \cline{2-7} 
                         
                         & \multirow{2}{*}{16}  & \multicolumn{1}{l|}{WNR}             & \multicolumn{1}{c|}{77.92} & \multicolumn{1}{l|}{26.66}  & \multicolumn{1}{l|}{24.55}  & \multicolumn{1}{l|}{34.02}       \\ 
                         &  & WNR + $R_n$ & \multicolumn{1}{c|}{73.77} & \multicolumn{1}{c|}{\textbf{42.71} ({\color{blue}{$\uparrow$ 16.05}})} & \multicolumn{1}{c|}{\textbf{42.71} ({\color{blue}{$\uparrow$ 18.16}})} & \textbf{50.63} ({\color{blue}{$\uparrow$ 16.61}})        \\ \cline{2-7}
                         
                         & \multirow{2}{*}{50}  & \multicolumn{1}{l|}{WNR}             & \multicolumn{1}{c|}{82.58} & \multicolumn{1}{l|}{11.36} & \multicolumn{1}{l|}{8.23} & \multicolumn{1}{l|}{17.34}      \\ 
                         & & WNR + $R_n$  & \multicolumn{1}{c|}{75.19} & \multicolumn{1}{c|}{33.12 ({\color{blue}{$\uparrow$ 21.76}})} & \multicolumn{1}{c|}{31.60 ({\color{blue}{$\uparrow$ 23.37}})} & 40.11 ({\color{blue}{$\uparrow$ 22.77}})           \\ \cline{1-7} 
                         
\multirow{8}{*}{Alexnet} & \multirow{2}{*}{1}     & \multicolumn{1}{l|}{WNR} & \multicolumn{1}{c|}{42.75} & \multicolumn{1}{l|}{10.20}  & \multicolumn{1}{l|}{8.72}  & \multicolumn{1}{l|}{15.80}        \\ 
                         &  & WNR + $R_n$ & \multicolumn{1}{c|}{56.08} & \multicolumn{1}{c|}{24.02 ({\color{blue}{$\uparrow$ 14.82}})} & \multicolumn{1}{c|}{23.13 ({\color{blue}{$\uparrow$ 14.41}})} & 30.55 ({\color{blue}{$\uparrow$ 14.75}})       \\ \cline{2-7}                       
                         & \multirow{2}{*}{2}  & \multicolumn{1}{l|}{WNR}             & \multicolumn{1}{c|}{50.17} & \multicolumn{1}{l|}{15.37}  & \multicolumn{1}{l|}{14.14}  & \multicolumn{1}{l|}{21.92}       \\ 
                         &  & WNR + $R_n$ & \multicolumn{1}{c|}{60.34} & \multicolumn{1}{c|}{32.89 ({\color{blue}{$\uparrow$ 17.52}})} & \multicolumn{1}{c|}{32.24 ({\color{blue}{$\uparrow$ 18.10}})} & 40.50 ({\color{blue}{$\uparrow$ 18.58}})       \\ \cline{2-7} 
                         & \multirow{2}{*}{4}  & \multicolumn{1}{l|}{WNR}             & \multicolumn{1}{c|}{59.14} & \multicolumn{1}{l|}{17.54}  & \multicolumn{1}{l|}{16.03}  & \multicolumn{1}{l|}{25.08}       \\ 
                         &  & WNR + $R_n$ & \multicolumn{1}{c|}{65.82} & \multicolumn{1}{c|}{31.00 ({\color{blue}{$\uparrow$ 13.46}})} & \multicolumn{1}{c|}{30.85 ({\color{blue}{$\uparrow$ 14.82}})} & 38.26 ({\color{blue}{$\uparrow$ 13.18}})       \\ \cline{2-7} 
                         
                         & \multirow{2}{*}{8}  & \multicolumn{1}{l|}{WNR}             & \multicolumn{1}{c|}{69.89} & \multicolumn{1}{l|}{19.65}  & \multicolumn{1}{l|}{18.30}  & \multicolumn{1}{l|}{27.73}       \\ 
                         &  & WNR + $R_n$ & \multicolumn{1}{c|}{70.37} & \multicolumn{1}{c|}{34.13 ({\color{blue}{$\uparrow$ 14.48}})} & \multicolumn{1}{c|}{33.61 ({\color{blue}{$\uparrow$ 15.31}})} & 41.08 ({\color{blue}{$\uparrow$ 13.35}})          \\ \cline{2-7} 
                         
                         & \multirow{2}{*}{16}  & \multicolumn{1}{l|}{WNR}             & \multicolumn{1}{c|}{77.92} & \multicolumn{1}{l|}{15.98} & \multicolumn{1}{l|}{14.04} & \multicolumn{1}{l|}{21.34}       \\ 
                         & & WNR + $R_n$  & \multicolumn{1}{c|}{73.77} & \multicolumn{1}{c|}{\textbf{33.31} ({\color{blue}{$\uparrow$ 17.33}})} & \multicolumn{1}{c|}{\textbf{31.72} ({\color{blue}{$\uparrow$ 17.68}})} & \textbf{40.56} ({\color{blue}{$\uparrow$ 19.22}})      \\ \cline{2-7} 

                         & \multirow{2}{*}{50}  & \multicolumn{1}{l|}{WNR}             & \multicolumn{1}{c|}{82.58} & \multicolumn{1}{l|}{5.58} & \multicolumn{1}{l|}{3.33} & \multicolumn{1}{l|}{10.44}       \\ 
                         & & WNR + $R_n$  & \multicolumn{1}{c|}{75.19} & \multicolumn{1}{c|}{19.65 ({\color{blue}{$\uparrow$ 14.07}})} & \multicolumn{1}{c|}{18.38({\color{blue}{$\uparrow$ 15.04}})} & 25.41 ({\color{blue}{$\uparrow$ 14.97}})      \\ \hline
\end{tabular}
}

\end{table*}

\section{Architecture details of Regenerator Network}

The Regenerator network consists of U-net-based~\cite{Ronneberger2015UNetCN} generator with five downsampling and upsampling layers. Each $i^{th}$ downsampling layer has a skip connection to $(n-i)^{th}$ upsampling layer that concatenates channels of $i^{th}$ layer with those at layer $n-i$, $n$ represents number of upsampling and downsampling layers (i.e. $n=5$). The number of channels in the network's input and output are the same as the image channels in the training dataset (surrogate data $S_d$ in our case). Each Downsampling layer first filters input through Leaky relu with negative $slope=0.2$, followed by convolution operation with $kernelsize=4$, $stride=2$, and $padding=1$. The number of output channels for layers $1$ to $5$ are $64$,$128$,$256$,$512$,$512$ respectively. Upsampling layers start with a Relu layer. Followed by transposed convolution. Each transposed convolution has $kernelsize=4$ , $padding=1$ and $stride=2$. The number of output channels for layers $1$ to $5$ are $1024$, $512$, $256$, $128$, and $3$ 
respectively.The output of convolution and deconvolution layers is normalized using instance normalization to avoid 'instance-specific mean and covariance shifts'. The output of the last upsampling layer is normalized using Tanh normalization to ensure output values lie in the range $[-1,1]$.

\begin{table*}[h]
\centering
\caption{Ablation on defense components of proposed DBMA and comparison with baseline. Both the components individually provide better defense than baseline. DBMA yields best performance when WNR and $R_n$ are used together.}
\label{tab:combination_of_wnr_rn}
\scalebox{0.85}{
\begin{tabular}{|c|c|cccc|}
\hline
\multirow{-1.5}{*}{\begin{tabular}[c]{@{}c@{}}Surrogate\\ model \\ (attacker)\end{tabular}} &
  \multirow{2}{*}{\begin{tabular}[c]{@{}c@{}} Defense\\Components \end{tabular}} &
  \multicolumn{4}{c|}{\begin{tabular}[c]{@{}c@{}}Black Box Model : Alexnet\\ Surrogate Model (defense): Resnet-18\end{tabular}} \\ \cline{3-6} 
 &
   &
  \multicolumn{1}{c|}{clean} &
  \multicolumn{1}{c|}{BIM} &
  \multicolumn{1}{c|}{PGD} &
  Auto Attack \\ \hline
\multirow{2}{*}{\begin{tabular}[c]{@{}c@{}}Alexnet-\\ half\end{tabular}} &
 \multicolumn{1}{l|}{Baseline}  &
  
  \multicolumn{1}{c|}{82.58} &
  \multicolumn{1}{l|}{7.02} &
  \multicolumn{1}{l|}{4.53} &
  \multicolumn{1}{l|}{11.65} \\ 
 &
  \multicolumn{1}{l|}{WNR}  &
  \multicolumn{1}{c|}{77.92} & \multicolumn{1}{c|}{{26.66} ({\color{blue}{$\uparrow$ 19.69}})} & \multicolumn{1}{c|}{{24.55} ({\color{blue}{$\uparrow$ 20.02}})} & {34.02} ({\color{blue}{$\uparrow$ 22.37}}) \\ 
  
 &
  \multicolumn{1}{l|}{$R_n$}  &
  \multicolumn{1}{c|}{77.03} & \multicolumn{1}{l|}{{29.40} ({\color{blue}{$\uparrow$ {22.38}}})} & \multicolumn{1}{c|}{{28.32} ({\color{blue}{$\uparrow$ 23.79}})} & {37.16} ({\color{blue}{$\uparrow$ 25.51}}) \\ 
  
   &
  \multicolumn{1}{l|}{WNR + $R_n$}  &
  \multicolumn{1}{c|}{73.77} & \multicolumn{1}{c|}{\textbf{42.71} ({\color{blue}{$\uparrow$ 35.69}})} & \multicolumn{1}{c|}{\textbf{42.71} ({\color{blue}{$\uparrow$ 38.18}})} & \textbf{50.63} ({\color{blue}{$\uparrow$ 38.98}}) \\ 
  \hline

\multirow{2}{*}{Alexnet} &
 \multicolumn{1}{l|}{Baseline} &
  \multicolumn{1}{c|}{82.58} &
  \multicolumn{1}{l|}{4.17} &
  \multicolumn{1}{l|}{2.19} &
  \multicolumn{1}{l|}{8.55} \\ 
 &
  \multicolumn{1}{l|}{WNR} &
  \multicolumn{1}{c|}{77.92} & \multicolumn{1}{c|}{{15.98} ({\color{blue}{$\uparrow$ 11.81}})} & \multicolumn{1}{c|}{{14.04} ({\color{blue}{$\uparrow$ 11.85}})} & {21.34} ({\color{blue}{$\uparrow$ 12.79}}) \\ 
  
   &
  \multicolumn{1}{l|}{$R_n$}  &
  \multicolumn{1}{c|}{77.03} & \multicolumn{1}{c|}{{16.52} ({\color{blue}{$\uparrow$ 12.35}})} & \multicolumn{1}{c|}{{15.09} ({\color{blue}{$\uparrow$ 12.9}})} & {22.40} ({\color{blue}{$\uparrow$  13.85}}) \\ 
  
   &
  \multicolumn{1}{l|}{WNR + $R_n$}  &
  \multicolumn{1}{c|}{73.77} & \multicolumn{1}{c|}{\textbf{33.31} ({\color{blue}{$\uparrow$ 29.14}})} & \multicolumn{1}{c|}{\textbf{31.72} ({\color{blue}{$\uparrow$ 29.53}})} & \textbf{40.56} ({\color{blue}{$\uparrow$ 32.01 }}) \\ 
  \hline

\end{tabular}
}
\end{table*}

\section{Importance of Regenerator Network}


To evaluate the effectiveness of proposed Regenerator network $R_n$ in our approach DBMA, we analyze the performance of DBMA with and without $R_{n}$ for different values of $k$ (i.e. $k$=$1$, $2$, $4$, $8$, $16$ (ours), $50$). 
    In Table~\ref{tab:rn_network}, we observe that across different $k$, appending $R_{n}$ to the WNR improves the adversarial accuracy against various attacks crafted using Alexnet-half and Alexnet by $\approx 13-18\%$. 
    Further, we observe a similar trend for clean accuracy, which also improves on adding $R_n$ to WNR. However, the improvement margin for clean accuracy gradually drops on increasing the value of coefficient percent $k$. For higher $k$'s (e.g., 16 , 50), there is a small drop in clean performance using $R_{n}$ 
    compared to the performance with only WNR but leads to significant increase in adversarial accuracy. This implies regenerator network enhances the output image of WNR to increase the adversarial accuracy.  In this process, for smaller values of $k$, it increases clean accuracy too, but for a large value of $k$, the decrease in clean accuracy is compensated by the increase in adversarial accuracy to achieve the best trade-off. Combining WNR with the regenerator network at our $\hat{k}$ (i.e., $k=16$) produces the best adversarial accuracy.

From Table~\ref{tab:combination_of_wnr_rn}, we observe that DBMA with only WNR improves adversarial accuracy with a small drop in clean accuracy compared to baseline. Similarly, with only regenerator network $R_n$, adversarial accuracy increases compared to the baseline. Also, $R_n$ performs better than WNR. However, using WNR and $R_n$ together in DBMA gives the best adversarial accuracy. Hence this demonstrates the importance of both the defense components (WNR and $R_n$) in our method DBMA. 

\section{Ablation on choice of surrogate architecture}
\label{sec_supp:choice_surr}

To better analyze the performance of DBMA against different combinations of defender surrogate ($S_m^d$) and attacker surrogate models ($S_m^a$), we perform experiments with different choices of surrogate models (i.e., Alexnet-half, Alexnet, and Resnet18). For all the experiments, Alexnet is used as the black box model. For Resnet18 as $S_m^d$, the wavelet coefficient selection module (WCSM) yields optimal $k$ ($\hat k$) as $16$. Similarly, for other choices of $S_m^d$ (i.e., Alexnet and Alexnet-half), we obtain $\hat k$ as $15$. This shows that the value of $\hat k$ is not much sensitive 
to the choice of architecture of the defender's surrogate model $S_m^d$. The results are reported in Table~\ref{tab:table_4}. 
\begin{table*}[htp]
\centering
\caption{Investigating the effect of surrogate model's architecture (for both defender and attacker) on the performance of our proposed approach (DBMA). Given defender's surrogate model, the attack is stronger if larger surrogate model is used by the attacker.}
\label{tab:table_4}
\scalebox{0.9}{
\begin{tabular}{|c|c|cccc|}
\hline
\multirow{2}{*}{\begin{tabular}[c]{@{}c@{}}Surrogate model \\ (defender)\end{tabular}} &
  \multirow{2}{*}{\begin{tabular}[c]{@{}c@{}}Surrogate model \\ (attacker)\end{tabular}} &
  \multicolumn{4}{c|}{Black Box Model : Alexnet} \\ \cline{3-6} 
             &              & \multicolumn{1}{c|}{clean} & \multicolumn{1}{c|}{BIM}   & \multicolumn{1}{c|}{PGD}   & Auto Attack \\ \hline

Resnet-18 & Alexnet-half & \multicolumn{1}{c|}{73.77} & \multicolumn{1}{c|}{\textbf{42.71}} & \multicolumn{1}{c|}{\textbf{42.71}} & \textbf{50.63} \\

Resnet-18 & Alexnet& \multicolumn{1}{c|}{73.77} & \multicolumn{1}{c|}{33.31} & \multicolumn{1}{c|}{31.72} & 40.56 \\
Resnet-18 & Resnet-18 & \multicolumn{1}{c|}{73.77} & \multicolumn{1}{c|}{22.48} & \multicolumn{1}{c|}{21.93} & 29.48 \\
Alexnet-half  & Alexnet-half & \multicolumn{1}{c|}{\textbf{74.94}} & \multicolumn{1}{c|}{38.98} & \multicolumn{1}{c|}{39.3} & 47.83 \\
Alexnet-half & Alexnet & \multicolumn{1}{c|}{74.94} & \multicolumn{1}{c|}{29.04} & \multicolumn{1}{c|}{27.58} & 35.37 \\
Alexnet-half & Resnet-18 & \multicolumn{1}{c|}{74.94} & \multicolumn{1}{c|}{20.72} & \multicolumn{1}{c|}{19.11} & 26.79 \\
Alexnet  & Alexnet-half & \multicolumn{1}{c|}{74.67} & \multicolumn{1}{c|}{40.08} & \multicolumn{1}{c|}{39.6} & 48.5 \\
Alexnet & Alexnet & \multicolumn{1}{c|}{74.67} & \multicolumn{1}{c|}{29.49} & \multicolumn{1}{c|}{28.63} & 36.93 \\
Alexnet & Resnet-18 & \multicolumn{1}{c|}{74.67} & \multicolumn{1}{c|}{21.51} & \multicolumn{1}{c|}{20.24} & 28.44 \\
 \hline
\end{tabular}
}

\end{table*}

We obtain the best performance for Resnet-18 as $S_m^d$ and Alexnet-half as $S_m^a$ ($1^{st}$ row), whereas the lowest  for Alexnet-half as $S_m^d$  and Resnet18 as $S_m^a$ ($6^{th}$ row). Further, on carefully observing the results, we deduce some key insights. Clean accuracy remains similar across different choices of surrogate models, but adversarial accuracy depends on the surrogate model of defender ($S_m^d$) and attacker ($S_m^a$). 
      
We observe that the bigger the network size, the more accurate the surrogate models. With accurate surrogate models, the gradients with respect to the input are better estimated. Thus better black-box attacks and defenses can be obtained using the bigger architectures for surrogate models. For better defense, $S_m^d$ should have a relatively higher capacity than $S_m^a$. This can be confirmed by rows 1, 4, and 7, where the defense becomes more effective on increasing the $S_m^d$'s capacity against various attacks using Alexnet-half as $S_m^a$. A similar trend is observed against the attacks crafted using Alexnet and Resnet18. For the other way around, i.e., when $S_m^a$ has relatively higher capacity than $S_m^d$, more powerful attacks can be crafted. This is evident from rows 1, 2, and 3, where stronger adversarial samples are obtained on increasing the capacity of $S_m^a$ for a given $S_m^d$. For instance, for Resnet18 as $S_m^d$, stronger attacks (lower adversarial accuracy) are obtained by using Resnet18 as $S_m^a$, followed by Alexnet and Alexnet-half. A similar pattern is observed on other choices of $S_m^d$.

\begin{table*}[!h]
\centering
\caption{Performance of our approach DBMA in defending the larger black-box network (i.e., Resnet34). Across various combinations, DBMA shows a consistent improvement against the different attacks with a small drop in the clean accuracy.}
\label{tab:resnet34}
\scalebox{0.85}{
\begin{tabular}{|l|l|l|l|l|l|l|}
\hline

\multirow{3}{*}{Method} &
  \multirow{3}{*}{\begin{tabular}[c]{@{}c@{}}Surrogate \\ Model\\ (Defender)\end{tabular}} &
  \multirow{3}{*}{\begin{tabular}[c]{@{}c@{}}Surrogate \\    Model\\ (Attacker)\end{tabular}} &
  \multicolumn{4}{c|}{\multirow{2}{*}{Black Box Model : Resnet34}} \\
    &          &          & \multicolumn{4}{c|}{}                                                                              \\ \cline{4-7} 
    &          &          & \multicolumn{1}{c|}{Clean} & \multicolumn{1}{c|}{BIM}   & \multicolumn{1}{c|}{PGD}   & Auto-Attack \\ \hline
Baseline & - & Resnet18 & 95.66 & \multicolumn{1}{l|}{3.11} & \multicolumn{1}{l|}{1.23} & \multicolumn{1}{l|}{11.82} \\
DBMA (\textbf{Ours}) & Alexnet & Resnet18 & 87.06 & \textbf{20.76}  & \textbf{17.33}  & \textbf{25.11}  \\ \hline
Baseline & - & Alexnet & 95.66 & \multicolumn{1}{l|}{21.36} & \multicolumn{1}{l|}{12.83} & \multicolumn{1}{l|}{27.85} \\
DBMA (\textbf{Ours}) & Resnet18 & Alexnet & 88.40 & \textbf{48.16}  & \textbf{44.80}  & \textbf{56.65}  \\ 
 \hline
\end{tabular} 
}

\end{table*}

\begin{table*}[!h]
\centering
\caption{Performance of our approach DBMA with fourier transform and wavelet transform based noise removal technique (FNR and WNR, respectively). WNR defense outperforms the FNR across different attacks. Also, WNR-based DBMA (WNR + $R_n$) yields more significant gains in performance on CIFAR10.}

\label{tab:fourier}
\scalebox{0.85}{
\begin{tabular}{|c|c|c|c|c|c|}
\hline
\multirow{-1.5}{*}{\begin{tabular}[c]{@{}c@{}}Surrogate\\ model \\ (attacker)\end{tabular}} &
  \multirow{-1.5}{*}{\begin{tabular}[c]{@{}c@{}}Method\end{tabular}} &
  \multicolumn{4}{c|}{\begin{tabular}[c]{@{}c@{}}Black Box Model : Alexnet\\ Surrogate Model (defense): Resnet-18\end{tabular}} \\ \cline{3-6} 
                         &        & \multicolumn{1}{c|}{clean} & \multicolumn{1}{c|}{BIM}   & \multicolumn{1}{c|}{PGD}   & \multicolumn{1}{c|}{Auto Attack} \\ \hline
                         
\multirow{5}{*}{\begin{tabular}[c]{@{}c@{}}Alexnet-\\ half\end{tabular}}
& \multicolumn{1}{l|}{Baseline} & \multicolumn{1}{c|}{82.58} & \multicolumn{1}{l|}{7.02} & \multicolumn{1}{l|}{4.53} & \multicolumn{1}{l|}{11.65} \\    \cline{2-6}   

& \multicolumn{1}{l|}{FNR} & 79.14 & \multicolumn{1}{l|}{13.30 ({\color{blue}{$\uparrow$ 6.28}})} & \multicolumn{1}{l|}{10.86 ({\color{blue}{$\uparrow$ 6.33}})} & \multicolumn{1}{l|}{20.47 ({\color{blue}{$\uparrow$ 8.82}})} \\
& \multicolumn{1}{l|}{FNR+ $R_n$} & 74.13 & 28.38 ({\color{blue}{$\uparrow$ 21.36}}) & 27.05 ({\color{blue}{$\uparrow$ 22.52}}) & 35.47 ({\color{blue}{$\uparrow$ 23.82}}) \\ \cline{2-6}  
& \multicolumn{1}{l|}{WNR} & 77.92 & 26.66 ({\color{blue}{$\uparrow$ 19.64}}) & 24.55 ({\color{blue}{$\uparrow$ 20.02}}) & 34.02 ({\color{blue}{$\uparrow$ 22.37}}) \\
& \multicolumn{1}{l|}{\begin{tabular}[l]{@{}l@{}}WNR + $R_n$\\(\textbf{Ours})\end{tabular}} & \textbf{73.77} & \textbf{42.71} ({\color{blue}{$\uparrow$ 35.69}}) & \textbf{42.71} ({\color{blue}{$\uparrow$ 38.18}}) & \textbf{50.63} ({\color{blue}{$\uparrow$ 38.98}}) \\ \hline
\multirow{5}{*}{Alexnet} 
& \multicolumn{1}{l|}{Baseline} & 82.58 & \multicolumn{1}{l|}{4.17} & \multicolumn{1}{l|}{2.19} & \multicolumn{1}{l|}{8.55} \\ \cline{2-6}
& \multicolumn{1}{l|}{FNR} & 79.14 & \multicolumn{1}{l|}{5.87 ({\color{blue}{$\uparrow$ 1.70}})} & \multicolumn{1}{l|}{4.03 ({\color{blue}{$\uparrow$ 1.84}})} & \multicolumn{1}{l|}{10.97 ({\color{blue}{$\uparrow$ 2.42}})} \\    
& \multicolumn{1}{l|}{FNR+ $R_n$} & 74.13 & \multicolumn{1}{l|}{19.24 ({\color{blue}{$\uparrow$ 15.07}})} & 17.88 ({\color{blue}{$\uparrow$ 15.69}}) & 24.55 ({\color{blue}{$\uparrow$ 16.00}}) \\ \cline{2-6} 
& \multicolumn{1}{l|}{WNR} & 77.92 & 15.98 ({\color{blue}{$\uparrow$ 11.81}}) & 14.04 ({\color{blue}{$\uparrow$ 11.85}}) & 21.34 ({\color{blue}{$\uparrow$ 12.79}}) \\
& \multicolumn{1}{l|}{\begin{tabular}[l]{@{}l@{}}WNR + $R_n$\\(\textbf{Ours})\end{tabular}} & \textbf{73.77} & \textbf{33.31} ({\color{blue}{$\uparrow$ 29.14}}) & \textbf{31.72} ({\color{blue}{$\uparrow$ 29.53}}) &\textbf{40.56} ({\color{blue}{$\uparrow$ 32.01}}) \\
 \hline
 
\end{tabular}
}

\end{table*}

\section{Our defense (DBMA) on larger black box model}
\label{sec_supp:larger_bb}

Throughout all our experiments, we used Alexnet as the black-box model $B_m$ . To check the consistency of our approach DBMA across different architecture, especially for bigger and high-capacity networks, 
we perform experiments using Resnet34 as black-box model and report corresponding results in Table~\ref{tab:resnet34}. With Resnet18 and Alexnet as the defender’s and attacker’s surrogate model ($S_m^d$ and $S_m^a$) respectively, we observe the improvement of $\approx 27-32\%$ in adversarial accuracy across attacks compared to baseline (rows $3^{rd}$ and $4^{th}$). With Alexnet as the defender’s surrogate and Resnet18 as the attacker’s surrogate model, we get an improvement of $\approx 17-23\%$ across attacks (rows $1^{st}$ and $2^{nd}$). As observed in previous experiments, compared to baseline, clean accuracy drops by $\approx7-8\%$. Overall, across different black box models, our proposed defense DBMA has obtained decent performance. Hence we conclude, DBMA is even effective 
on bigger black box architectures. 

\section{Comparison with Fourier Transform based Noise removal}

Apart from the wavelet transformations, some recent works utilised the fourier transformations to remove the adversarial noise from the adversarial images, and further found it to be effective in denoising ~\cite{Yin2019AFP}. In this section, we do an ablation on our choice of Wavelet-based Noise Remover (WNR) over the other possible choice of fourier-based Noise Remover (FNR). 

As observed in recent works~\cite{Yin2019AFP}, adversarial attack affects the high-frequency components more than low-frequency components. Therefore, In FNR we apply a low pass filter on an image with threshold radius $\hat{r}$. Similar to WCSM, we compute $LCR_C$ , $LCR_A$, $LCR$ and $ROC$ for different values of $r$. The optimal value of $r$ (i.e., $\hat{r}$) is selected at which $ROC$ starts saturating ($\hat{r}=11$). In Table~\ref{tab:fourier}, we observe, compared to baseline, Fourier-based DBMA gives an improvement of $\approx 21-34\%$ in adversarial accuracy across attacks
for Alexnet half (rows 1 and 3).
Compared to Fourier-based DBMA, the results using our wavelet-based DBMA are significantly better in terms of adversarial accuracy (rows 1 and 5) with similar clean performance. Similarly, for Alexnet, Wavelet-based DBMA gives $\approx 16-24\%$ better adversarial accuracy compared to Fourier based
DBMA across all attacks (rows 8 and 10). Hence, our wavelet-based DBMA is more robust across different adversarial attacks than Fourier-based DBMA.

\section{Algorithm}

\begin{algorithm}
\small
 \caption{Algorithm for our proposed method (DBMA)}
 \begin{algorithmic}[1]
 \renewcommand{\algorithmicrequire}{\textbf{Input:}}
 \renewcommand{\algorithmicensure}{\textbf{Output:}}
 \REQUIRE Black box model $B_{m}$, max coefficients $k^{max}$
 \ENSURE $\hat{B}_m$
 \\ {\underline{\textbf{Step 1: \textit{\texttt{Model Stealing}}}}} \vspace{1mm}
 \STATE Surrogate model $S_{m}$, Synthetic data $S_{d}$ $\leftarrow$ Model Stealing on $B_m$
 \\ \underline{\textbf{Step 2: \textit{\texttt{Wavelet Coefficient Selection Module}}}}
 \STATE Obtain adversarial samples ($S_{da}$) corresponding to $S_{d}$ using adversarial attack on $S_{m}$
 \FOR{$k=1:k^{max}$:}
 \STATE $\bar{S}_{da}^{k} \leftarrow$ WNR($S_{da}$,$k$)
 \STATE $N_{flips} = 0$
 \FOR{$i = 1 : (|\bar{S}_{da}^{k}|)$}
 \STATE $\bar{x}_{sa}^{i}$ $\leftarrow$ $\bar{S}_{da}^{k}[i]$ \hfill \COMMENT{$i^{th}$ element of $\bar{S}_{da}^{k}$}
 \STATE ${x}_{s}^{i}$ $\leftarrow$ ${S}_{d}[i]$ \hfill
 \COMMENT{$i^{th}$ element of $S_{d}$}
 \IF{$label(B_{m}(\bar{x}_{sa}^{i})) \neq label(B_{m}(x_{s}^i))$}
 \STATE $N_{flips} = N_{flips} +1$
 \ENDIF
 \ENDFOR
 \STATE $LFR^k = N_{flips} / |{S}_{d}|$
 \ENDFOR
 \STATE $\hat{k} = \underset{k}{\mathrm{argmin}} \hspace{0.1in}LFR^k$
 \\ \underline{\textbf{Step 3: \textit{\texttt{Training Regenerator network} ($R_n$)}}} \vspace{1mm}
 \STATE $\bar{S}_{d}^{\hat{k}} \leftarrow$ WNR($S_{d}$,$\hat{k}$) \vspace{0.5mm}
 \STATE $\bar{S}_{da}^{\hat{k}} \leftarrow$ WNR($S_{da}$,$\hat{k}$)
 \STATE Initialize $R_n^{\theta}$ 
 \FOR{$epoch < MaxEpoch$}
 \FOR{$i = 1 : (|{S}_{d}|)$}
 \STATE ${x}_{s}^{i}$ $\leftarrow$ ${S}_{d}[i]$ \hfill
 \COMMENT{$i^{th}$ element of $S_{d}$}
 \STATE $\bar{x}_{s}^{i}$ $\leftarrow$  $\bar{S}_{d}^{\hat{k}}[i]$ \hfill \COMMENT{$i^{th}$ element of $\bar{S}_{d}^{\hat{k}}$}
 \STATE $\bar{x}_{sa}^{i}$ $\leftarrow$ $\bar{S}_{da}^{\hat{k}}[i]$ \hfill \COMMENT{$i^{th}$ element of $\bar{S}_{da}^{\hat{k}}$}
 \STATE $L_{cs} = CS(S_m(R_n(\bar{x}_{s}^{i})),  S_m(x_{s}^{i}))$ \hfill \COMMENT{$CS$ is cosine similarity}
 \STATE $L_{kl} = KL(soft(S_m(R_n(\bar{x}_{sa}^{i}))), soft(S_m(R_n(\bar{x}_{s}^{i}))))$ \hfill \COMMENT{$KL$ is KL divergence}
 \STATE $L_{sc} = \left\lVert R_n(\bar{x}_{s}^{i}) - x_{s}^{i} \right\rVert_{1} + \left\lVert R_n(\bar{x}_{sa}^{i}) - x_{s}^{i} \right\rVert_{1}$
 \STATE $L(R_n^{\theta}) = -\lambda_1 L_{cs} + \lambda_2 L_{kl} + \lambda_3 L_{sc}$ 
 \STATE Update $R_n^{\theta}$ by minimizing $L(R_n^{\theta})$ using Adam Optimizer
 \ENDFOR
 \ENDFOR
 \STATE $\hat{B}_m$ $\leftarrow$ concatenate($WNR(.,\hat{k})$, $R_n^{\theta^{*}}$, $B_{m}$) \hfill \COMMENT{The black box model $\hat{B}_m$ is used by attacker}  
 \RETURN $\hat{B}_m$ 
 \end{algorithmic} 
 \end{algorithm}

\newpage
\section{Visualization}
\begin{figure}[h]
\centering
\centerline{\includegraphics[width=0.5\textwidth]{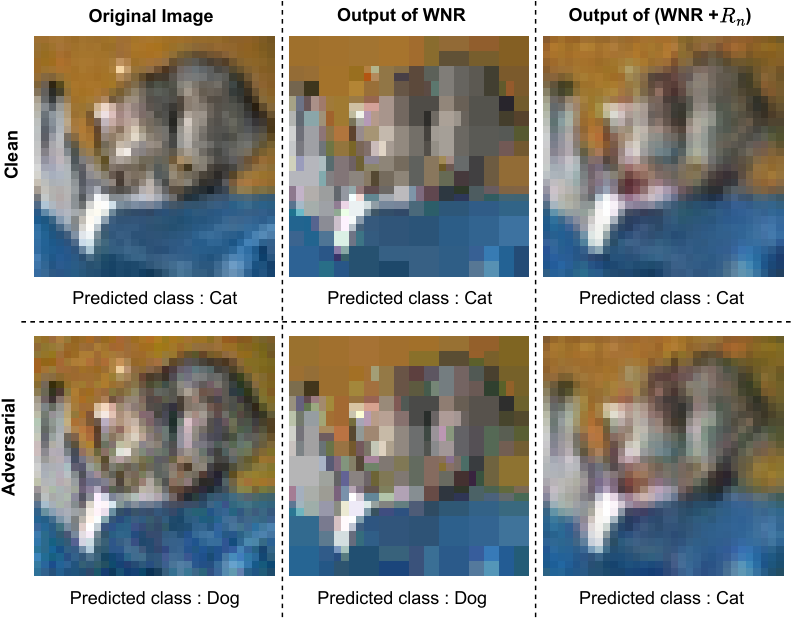}}
\caption{Visualization of images: The top row indicates input as clean image and bottom row corresponds to adversarial image.  
The predictions obtained by the black-box network on inputs: (a) Original clean image  (b) Output of wavelet noise remover on clean image (c) Output of WNR with regenerator network ($R_{n}$) on clean image (d) Original adversarial image  (e) Output of wavelet noise remover (WNR) on adversarial image (f) Output of WNR with regenerator network ($R_n$) on adversarial Image. Here, the ground truth class is Cat. Our method (DBMA) produces correct output using regenerated image as input.
}
\label{fig: label_consistency_rate}
\end{figure}
\begin{table*}[h]
\centering
\caption{Attack parameters for different adversarial attacks: BIM, PGD and Auto Attack}
\label{tab:table_5}
\scalebox{0.9}{
\begin{tabular}{|c|c|ccc|}
\hline
\multirow{2}{*}{Dataset} & \multirow{2}{*}{Attack Parameters} & \multicolumn{3}{c|}{Adversarial Attacks}                                          \\ \cline{3-5} 
                         &                                    & \multicolumn{1}{c|}{BIM}   & \multicolumn{1}{c|}{PGD}   & Auto Attack \\ \hline
\multirow{3}{*}{CIFAR-$10$} & $\epsilon$                            & \multicolumn{1}{c|}{8/255} & \multicolumn{1}{c|}{8/255} & 8/255       \\ \cline{2-5} 
 & $\epsilon_{step}$ & \multicolumn{1}{c|}{\begin{tabular}[c]{@{}c@{}}0.00156 \\ ($\epsilon$/no of iterations)\end{tabular}} & \multicolumn{1}{c|}{2/255} & -- \\ \cline{2-5} 
                         & no of iterations                              & \multicolumn{1}{c|}{20}    & \multicolumn{1}{c|}{20}    & --          \\ \hline
\multirow{3}{*}{SVHN}    & $\epsilon$                            & \multicolumn{1}{c|}{4/255} & \multicolumn{1}{c|}{4/255} & 4/255       \\ \cline{2-5} 
                         & $\epsilon_{step}$                              & \multicolumn{1}{c|}{2/255} & \multicolumn{1}{c|}{2/255} & --          \\ \cline{2-5} 
                         & no of iterations                              & \multicolumn{1}{c|}{20}    & \multicolumn{1}{c|}{20}    & --          \\ \hline
\end{tabular}
}

\end{table*}
\newpage

\section{Adversarial Attack Parameters and Training Details}
We evaluate the performance of black box model ($B_m$) on three adversarial attacks, PGD, BIM and Auto Attack. Parameters used for each attack are summarized in  Table~\ref{tab:table_5}. 

\textbf{Training details of Regenerator Network} ($R_n$): The regenerator network is trained with Adam optimizer with learning rate of $0.0002$ for 300 epochs. Learning rate is decayed using linear scheduler, where we keep the learning rate fixed for 100 epochs and then linearly decay the rate to zero. Batch size is set to $128$.


\newpage

\newpage
\end{document}